\pdfoutput=1
\documentclass[11pt]{article}

\usepackage{EMNLP2023}

\usepackage{times}
\usepackage{latexsym}

\usepackage[T1]{fontenc}
\usepackage[utf8]{inputenc}

\usepackage{microtype}

\usepackage{inconsolata}

\usepackage{adjustbox}
\usepackage{multirow}
\usepackage{multicol}
\usepackage{booktabs}
\usepackage{setspace}
\usepackage{hyperref}
\usepackage{threeparttable}
\usepackage{xcolor}
\usepackage{colortbl}
\usepackage{float}
\usepackage{enumitem}
\definecolor{r1}{RGB}{48,182,86}
\definecolor{r2}{RGB}{100,209,138}
\definecolor{r3}{RGB}{168,233,191}
\definecolor{r4}{RGB}{204,245,208}
\definecolor{r5}{RGB}{235,254,236}
\definecolor{sota}{RGB}{215,215,215}
\definecolor{sota_text}{RGB}{185,185,185}
\newcommand{\method}{{BioT5}}

\newcommand{\bom}{$\langle$bom$\rangle$}
\newcommand{\eom}{$\langle$eom$\rangle$}
\newcommand{\bop}{$\langle$bom$\rangle$}
\newcommand{\eop}{$\langle$eom$\rangle$}
\newcommand{\dataset}{$\langle$\texttt{Dataset}$\rangle$}
\newcommand{\selfies}{$\langle$\texttt{SELFIES}$\rangle$}
\newcommand{\fasta}{$\langle$\texttt{FASTA}$\rangle$}
\newcommand{\text}{$\langle$\texttt{Text Description}$\rangle$}
\title{BioT5: Enriching Cross-modal Integration in Biology with Chemical Knowledge and Natural Language Associations}

\author{
    Qizhi Pei\textsuperscript{1,5}, 
    Wei Zhang\textsuperscript{2}, 
    Jinhua Zhu\textsuperscript{2}, 
    Kehan Wu\textsuperscript{2},
    Kaiyuan Gao\textsuperscript{3}, \\
    {\bf Lijun Wu\textsuperscript{4}$^{\ast}$},
    {\bf Yingce Xia\textsuperscript{4}$^{\ast}$},
    {\bf Rui Yan\textsuperscript{1,6}\thanks{\ \ Corresponding authors: Lijun Wu (\url{lijuwu@microsoft.com}), Yingce Xia (\url{yinxia@microsoft.com}), and Rui Yan (\url{ruiyan@ruc.edu.cn})}} \\
    \textsuperscript{1}Gaoling School of Artificial Intelligence, Renmin University of China \\
    \textsuperscript{2}University of Science and Technology of China \\
    \textsuperscript{3}Huazhong University of Science and Technology \quad
    \textsuperscript{4}Microsoft Research\\
    \textsuperscript{5}Engineering Research Center of Next-Generation Intelligent Search\\ and Recommendation, Ministry of Education \\
    \textsuperscript{6}Beijing Key Laboratory of Big Data Management and Analysis Methods \\
    \texttt{\{qizhipei,ruiyan\}@ruc.edu.cn} \\
    \texttt{\{weizhang\_cs,teslazhu,wu\_2018\}@mail.ustc.edu.cn} \\
    \texttt{im\_kai@hust.edu.cn} \quad
    \texttt{\{lijuwu,yinxia\}@microsoft.com} 
}

\begin{document}
\maketitle
\begin{abstract}
Recent advancements in biological research leverage the integration of molecules, proteins, and natural language to enhance drug discovery. However, current models exhibit several limitations, such as the generation of invalid molecular SMILES, underutilization of contextual information, and equal treatment of structured and unstructured knowledge. To address these issues, we propose BioT5, a comprehensive pre-training framework that enriches cross-modal integration in biology with chemical knowledge and natural language associations. BioT5 utilizes SELFIES for 100\% robust molecular representations and extracts knowledge from the surrounding context of bio-entities in unstructured biological literature. Furthermore, BioT5 distinguishes between structured and unstructured knowledge, leading to more effective utilization of information. After fine-tuning, BioT5 shows superior performance across a wide range of tasks, demonstrating its strong capability of capturing underlying relations and properties of bio-entities. Our code is available at \url{https://github.com/QizhiPei/BioT5}.
\end{abstract}

\section{Introduction}
\begin{figure}
    \centering
    \includegraphics[width=\linewidth]{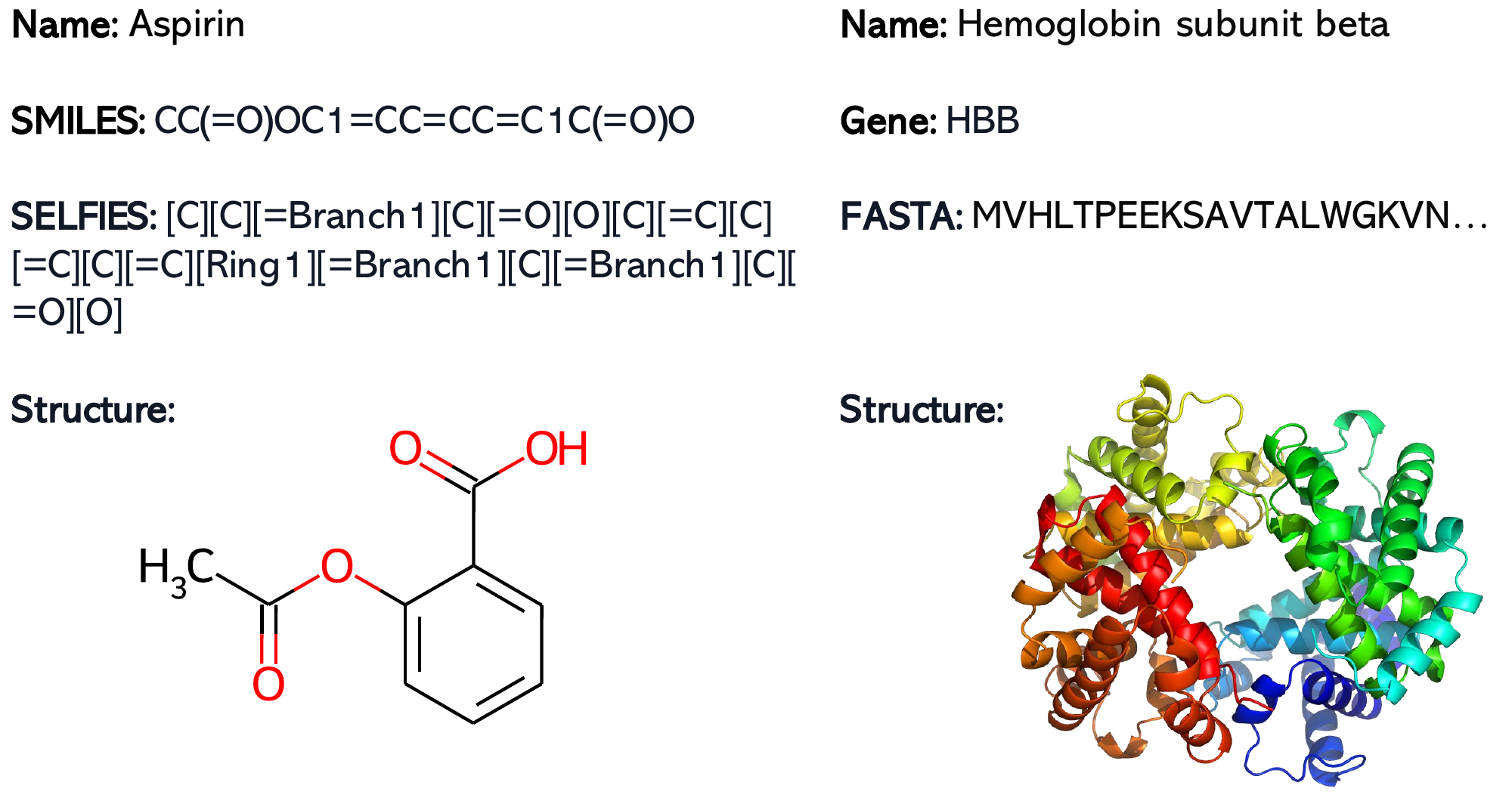}
    \caption{Representations of molecule and protein. Molecule can be represented by its name, bio-sequence (SMILES and SELFIES), and 2D graph structure. Protein can be represented by its name, corresponding gene name, bio-sequence (FASTA), and 3D structure.}
    \label{fig:molecule_protein_rep}
\end{figure}
Molecules and proteins are two essential bio-entities in drug discovery~\citep{dara2022machine}. 
% \footnote{In this paper, the term ``molecule'' refers to small molecule compound or micromolecule consisting of two or more atoms that are chemically bonded together, while ``protein'' represents biological macromolecule composed of amino acids.}
Small molecule drugs have been the cornerstone of the pharmaceutical industry for nearly a century, owing to their unique advantages such as oral availability, diverse modes of action, etc \citep{future_smallM}. Proteins serve as the foundation of life science, functioning as drug targets or crucial elements in disease pathways.
As illustrated in Figure~\ref{fig:molecule_protein_rep}, both molecules and proteins can be represented using sequences. 
A molecule can be depicted by a SMILES sequence~\citep{weininger1988smiles,weininger1989smiles}, which is derived by traversing the molecular graph through depth-first search and applying specific branching rules. 
A protein can be represented by a FASTA sequence~\citep{lipman1985rapid,pearson1988improved}, which outlines the amino acids in a protein.
The sequential formats of molecules and proteins facilitate the application of Transformer models~\citep{vaswani2017attention} and pre-training techniques~\citep{liu2019roberta,radford2019language} from natural language processing (NLP) to the biomedical field. 
Chemberta~\citep{chithrananda2020chemberta} and ESM~\citep{rives2021biological,lin2022language} apply masked language modeling to molecular SMILES and protein FASTA respectively, while MolGPT~\citep{DBLP:journals/jcisd/BagalAVP22} and ProtGPT2~\citep{ferruz2022protgpt2} leverage GPT-style models for molecular and protein generation.

Scientific literature~\citep{DBLP:conf/emnlp/BeltagyLC19,canese2013pubmed} and biological databases~\citep{kim2023pubchem,boutet2007uniprotkb} serve as knowledge repositories of molecules and proteins.
These resources detail properties, experimental results, and interactions between various bio-entities, which cannot be explicitly inferred from molecular or protein sequences alone.
Consequently, a recent trend involves jointly modeling text along with molecules and proteins, allowing the textual descriptions to enhance molecular and protein representations. 
MolT5~\citep{DBLP:conf/emnlp/EdwardsLRHCJ22} adopts the T5~\citep{raffel2020exploring} framework to molecular SMILES and biomedical literature. 
MolXPT~\citep{liu2023molxpt} and Galactica~\citep{taylor2022galactica} are GPT models trained on text and bio-entities, such as SMILES and FASTA sequences.
DeepEIK~\citep{luo2023empowering} fuses the encoded features from multi-modal inputs using attention~\citep{vaswani2017attention} mechanism.
Despite their success, there is still significant room for improvement: (i) Prior work often relies on SMILES to represent molecules. However, addressing the issue of generating invalid SMILES remains a challenge to overcome~\citep{DBLP:conf/emnlp/EdwardsLRHCJ22,li2023empowering}.
(ii) The contextual information surrounding molecular or protein names could offer valuable insights for understanding the interactions and properties of bio-entities. Developing effective methods to leverage this information merits further attention. 
(iii) Existing research tends to treat structured data (e.g., molecule-text pairs from databases) and unstructured data (e.g., text sequences in literature) equally. However, structured data could be utilized more effectively to further enhance overall performance.

To address the above challenges, in this paper, we introduce \textbf{\method}, a comprehensive pre-training framework encompassing text, molecules, and proteins. \method{} leverages SELFIES \cite{krenn2020self}  to represent small molecules since its advantage over SMILES is that SELFIES offers a more robust and error-tolerant molecular representation, eliminating issues of illegitimate structures often encountered with SMILES. There are mainly two steps for \method{} pre-training:

(1) {\em Data collection \& processing}: We gather text, molecule, and protein data, as well as existing databases containing molecule-text parallel data and protein-text parallel data. For the text data (PubMed) from the biological domain, we employ named entity recognition and entity linking to extract molecular and protein mentions, replacing them with the corresponding SELFIES or FASTA sequences. Following~\citet{liu2023molxpt}, we refer to such data as ``wrapped'' text. Text tokens, FASTA sequences, and SELFIES are tokenized independently (see Section~\ref{sec:sep_token_emb} for more details).

(2) {\em Model training}: \method{} utilizes a shared encoder and a shared decoder to process various modalities. The standard T5 employs the ``recover masked spans'' objective, wherein each masked span and its corresponding part share the same sentinel token. We refer to the aforementioned training objective function as the ``T5 objective'' for simplicity. There are three types of pre-training tasks: (i) Applying the standard T5 objective to molecule SELFIES, protein FASTA, and general text independently, ensuring that the model possesses capabilities in each modality. (ii) Applying the T5 objective to wrapped text from the biological domain, where all text, FASTA, and SELFIES tokens can be masked and recovered. (iii) For the structured molecule-text data, we introduce a translation objective. Specifically, \method{} is trained to translate molecule SELFIES to the corresponding description and vice versa. Likewise, the translation objective is applied to protein-text data. 

After pre-training, we fine-tune the obtained \method{} on $15$ tasks covering molecule and protein property prediction, drug-target interaction prediction, protein-protein interaction prediction, molecule captioning, and text-based molecule generation. \method{} achieves state-of-the-art performances on $10$ tasks and exhibits results comparable to domain-specific large models on $5$ tasks, demonstrating the superior ability of our proposed method. 
\method{} model establishes a promising avenue for the integration of chemical knowledge and natural language associations to augment the current understanding of biological systems.

\section{Related Work}
In this section, we briefly review related work about cross-modal models in biology and representations of molecule and protein.

\subsection{Cross-modal Models in Biology}
\label{sec:cross_modal}
Language models in the biology field have gained considerable attention.
Among these, BioBERT~\citep{lee2020biobert} and BioGPT~\citep{luo2022biogpt}, which are pre-trained on scientific corpora, have been particularly successful in effectively understanding scientific texts.
More recently, cross-modal models focusing on jointly modeling text with bio-sequences have emerged.
They can be categorized into the following three groups.

\noindent{\textbf{Cross Text-molecule Modalities}}
MolT5~\citep{DBLP:conf/emnlp/EdwardsLRHCJ22} is a T5~\citep{raffel2020exploring}-based model, which is jointly trained on molecule SMILES and general text corpus.
MoSu~\citep{su2022molecular} is trained on molecular graphs and related textual data using contrastive learning.
MolXPT~\cite{liu2023molxpt} is a GPT~\citep{radford2018improving}-based model pre-trained on molecule SMILES, biomedical text, and wrapped text.
Different from \method{}, these models all use SMILES to represent molecules, which leads to validity issues when generating molecules.

\noindent{\textbf{Cross Text-protein Modalities}}
ProteinDT~\citep{liu2023text} is a multi-modal framework that uses semantically-related text for protein design.
BioTranslator~\citep{xu2023multilingual} is a cross-modal translation system specifically designed for annotating biological instances, such as gene expression vectors, protein networks, and protein sequences, based on user-written text. 

\noindent{\textbf{Cross Three or More Biology Modalities}}
Galactica~\citep{taylor2022galactica} is a general GPT-based large language model trained on various scientific domains, including scientific paper corpus, knowledge bases (e.g., PubChem~\citep{kim2023pubchem} molecules, UniProt~\citep{uniprot2023uniprot} protein), codes, and other sources.
DeepEIK~\cite{luo2023empowering} fuses the feature from multi-modal inputs (drugs, proteins, and text). Then attention~\citep{vaswani2017attention} mechanism is adopted to do textual information denoising and heterogeneous features integration. 

Our work differs from previous studies in several ways:
(1) we primarily focus on two biological modalities—molecule, protein-with text serving as a knowledge base and bridge to enrich the underlying relations and properties in the molecule and protein domains;
(2) we use multi-task pre-training to model the connections between these three modalities in a more comprehensive manner.
(3) we use SELFIES instead of SMILES to represent molecules, which is more robust and resolves the validity issue in molecule generation tasks.
\begin{figure*}[t]
    \centering
    \includegraphics[width=0.9\linewidth]{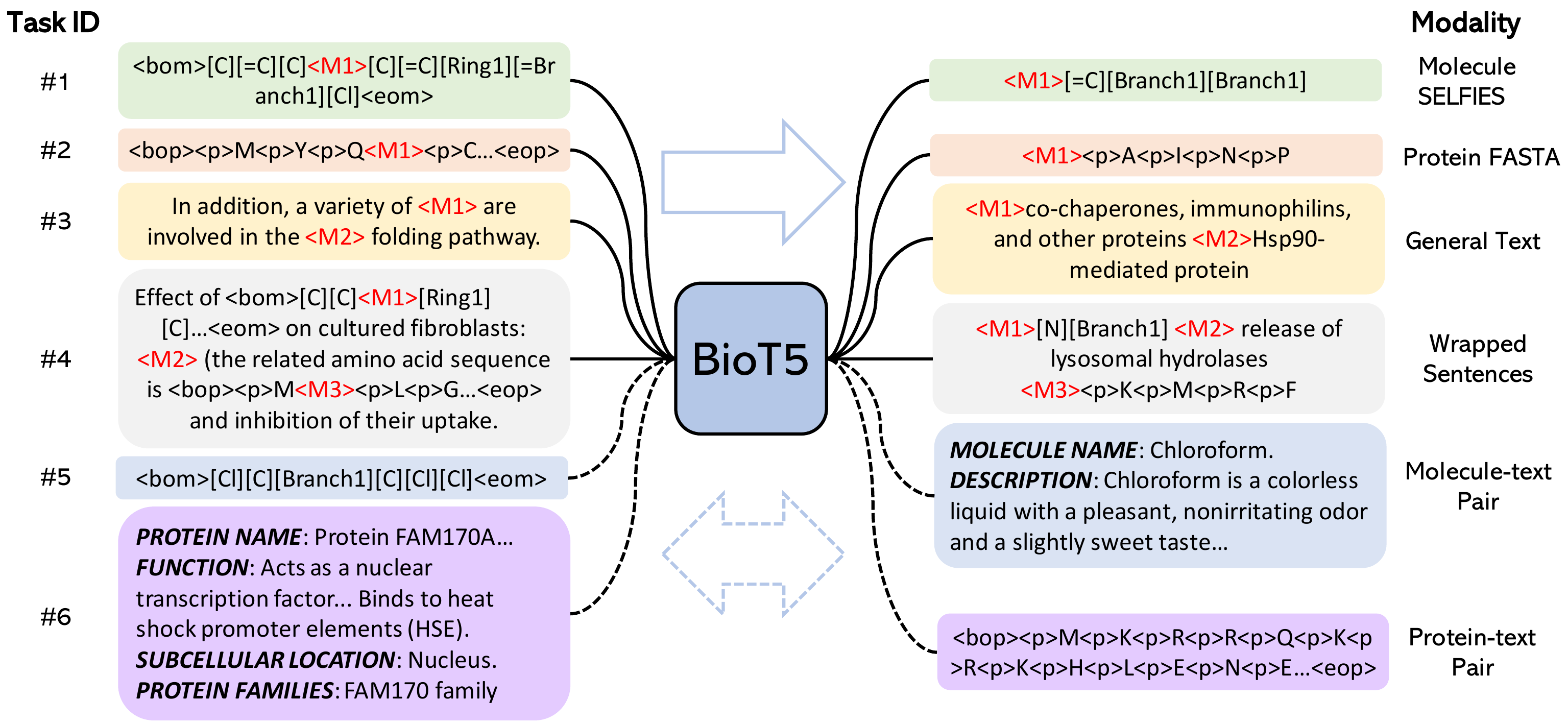}
    \caption{Overview of \method{} pre-training. The solid line refers to the ``T5 objective'', which aims to reconstruct the original unmasked input.
    Each consecutive span of masked tokens is replaced with a sentinel token, depicted as \texttt{<M1>}, \texttt{<M2>}, and \texttt{<M3>}.
    We apply this objective to molecule SELFIES (task \#1), protein FASTA (task \#2), general text (task \#3), and wrapped text (task \#4). 
    The dashed line represents the bidirectional translation between bio-sequences and structured text description (task \#5 and \#6).}
    \label{fig:pipeline}
\end{figure*}
\subsection{Representations of Molecule and Protein}
\label{sec:mol_pro_rep}
\noindent{\textbf{Molecule Representation}}
The representation and modeling of molecules have long been a challenge in bioinformatics.
There are many methods to represent a molecule: name, fingerprint~\citep{rogers2010extended}, SMILES~\citep{weininger1988smiles,weininger1989smiles}, InChl~\citep{heller2013inchi}, DeepSMILES~\citep{o2018deepsmiles}, SELFIES~\citep{krenn2020self}, 2D molecular graph, etc.
SMILES (Simplified Molecular-Input Line-Entry System), a compact and textual representation of the molecular structure, is the most common method. It employs a sequence of characters to encode atoms, bonds, and other molecular features.
However, SMILES has several drawbacks~\citep{krenn2022selfies}, such as the lack of syntactic and semantic robustness, which significantly affects the validity of molecules generated by deep learning models~\citep{DBLP:conf/emnlp/EdwardsLRHCJ22}.
To address this issue, SELFIES (Self-referencing Embedded Strings) is introduced as a 100\% robust molecular string representation~\citep{krenn2020self}.
Every permutation of symbols within the SELFIES alphabet invariably generates a chemically valid molecular structure, ensuring that each SELFIES corresponds to a valid molecule. 
Unlike existing works introduced in Section~\ref{sec:cross_modal} that use SMILES for molecule representation, we employ SELFIES with separate encoding in \method{} to achieve 100\% validity in downstream molecule generation tasks.

\noindent{\textbf{Protein Representation}}
Protein can also be represented in various ways, such as by its name, corresponding gene name, FASTA format, or 3D geometric structure.
The FASTA format is a common choice for encoding protein sequences, which uses single-letter codes to represent the $20$ different amino acids. 
In \method, we also employ FASTA format for protein representation.

Unlike~\citet{DBLP:conf/emnlp/EdwardsLRHCJ22} and~\citet{taylor2022galactica} that share the dictionary between bio-sequence tokens and nature language tokens, \method{} uses a separate dictionary and biology-specific tokenization to explicitly distinguish biological modalities. 
We give further analysis of this in Section~\ref{sec:sep_token_emb}.

\section{BioT5}
The overview of the \method{} pre-training is illustrated in Figure~\ref{fig:pipeline}.
We combine data from different modalities to perform multi-task pre-training.

\subsection{Pre-training Corpus}
As shown in Figure~\ref{fig:pipeline}, the pre-training corpus of \method{} is categorized into three classes:
(1) {\em Single-modal data}, including molecule SELFIES, protein FASTA, and general text.
For small molecules, we use the ZINC20~\citep{irwin2020zinc20} dataset and convert SMILES to SELFIES.
For protein FASTA, we randomly sample proteins from the Uniref50~\citep{suzek2007uniref} dataset, filtering out proteins exceeding a specified length, resulting in a collection of $27$M proteins
For general text, we use the ``Colossal Clean Crawled Corpus'' (C4) dataset~\citep{raffel2020exploring}.
(2) {\em Wrapped text}, where molecule names are replaced with their corresponding SELFIES and gene names are appended with related protein FASTA. 
We use $33$M PubMed articles~\citep{canese2013pubmed} and apply BERN2~\citep{sung2022bern2} for named entity recognition.
The scientific sentences which are not replaced or appended by bio-sequences are remained as a supplement to general text.
The detailed process is depicted in Figure~\ref{fig:ner} and discussed in Appendix~\ref{sec:ner_el}.
(3) {\em Molecule-description pairs} and {\em protein-description pairs}.
For molecule-text data, we collect $339$K molecule SELFIES along with their corresponding names and descriptions from PubChem~\citep{kim2019pug}, excluding all molecules present in the downstream ChEBI-20 dataset~\citep{DBLP:conf/emnlp/EdwardsLRHCJ22} to avoid potential data leakage.
For protein-text data, we obtain $569$K protein FASTA-description pairs from Swiss-Prot~\citep{boutet2007uniprotkb}, which contains high-quality annotations of various protein properties. 
Details are left in Appendix~\ref{sec:special_tokens}.

\subsection{Separate Tokenization and Embedding}
\label{sec:sep_token_emb}
\begin{figure}[t]
    \centering
    \includegraphics[width=\linewidth]{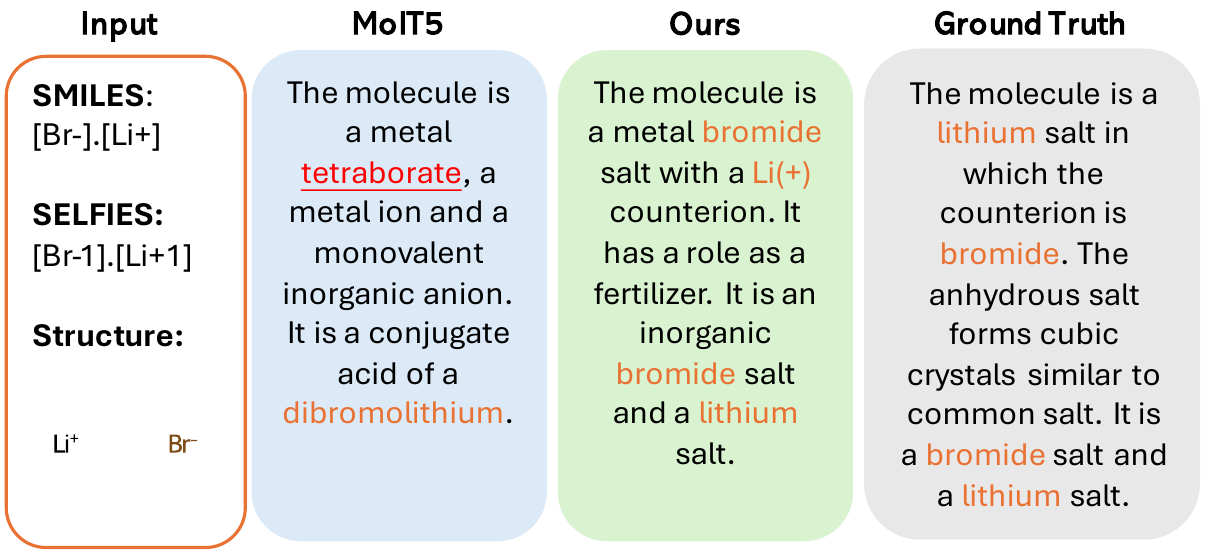}
    \caption{Case for tokenization. MolT5 processes ``Br''(bromine atom) as ``B'' (boron atom) and ``r'', resulting in incorrect descriptions including tetraborate (related to ``B''). \method{} retains the chemically meaningful group ``[Br-1]'' as a complete token, thereby producing the correct output.}
    \label{fig:case_tokenization}
    % \vspace{-0.5cm}
\end{figure}
In most previous works, the representation of molecules and proteins has not been modeled with sufficient attention to detail.
MolT5~\citep{DBLP:conf/emnlp/EdwardsLRHCJ22} employs the same dictionary as the original T5, as it starts pre-training from the original T5 checkpoint.
The original T5 dictionary is derived from nature language using SentencePiece~\citep{DBLP:conf/emnlp/KudoR18}.
However, directly utilizing this dictionary for molecule SMILES is suboptimal, as some chemically meaningful tokens, such as functional groups or complete atoms, will be tokenized inaccurately.
For example, in the molecule depicted in Figure~\ref{fig:case_tokenization}, the bromine atom, symbolized as ``Br'' in SMILES, is tokenized as ``B'' (a boron atom) and ``r'' by MolT5.
Consequently, MolT5 incorrectly characterizes this molecule as both dibromolit (related to ``Br'') and tetraborate (related to ``B'').
The character-based tokenization of Galactica~\citep{taylor2022galactica} suffers the same issue.

In addition to the tokenization method, sharing token embeddings for different modalities~\cite{DBLP:conf/emnlp/EdwardsLRHCJ22,taylor2022galactica} is also questionable.
In multilingual tasks, shared embeddings allow models to accurately represent the meanings of borrowed words and cognates, which retain their original meanings across languages.
However, molecules, proteins, and text represent entirely distinct languages.
The same token within these three different modalities carries different semantic meanings.
For example, the token ``C'' signifies character C in nature language, the carbon atom in molecules, and cysteine (one of the $20$ amino acids) in proteins. 
Studies by~\citet{DBLP:conf/emnlp/BeltagyLC19} and~\citet{gu2021domain} further emphasize the significance of domain-specific vocabulary.

To address the issues mentioned above, we employ separate vocabularies for molecule, protein, and text.
In \method{}, molecule is represented by SELFIES string, where each chemical meaningful atom group is enclosed within brackets and tokenized as a SELFIES token.
For example, \texttt{[C][=C][Br]}$\rightarrow$\texttt{[C],[=C],[Br]}.
For protein, to differentiate amino acids with capital letters in text, we introduce a special prefix \texttt{<p>} for each amino acid. For example, \texttt{<p>M<p>K<p>R}$\rightarrow$\texttt{<p>M,<p>K,<p>R}.
For text, we use the same dictionary as the original T5.
Through this, we explicitly distinguish the semantic space of different modalities, which maintains the inherent integrity of each unique modality and prevents the model from conflating meanings across modalities.

\subsection{Model and Training}
\noindent{\textbf{Model architecture}}
\method{} employs the same architecture as T5 models~\citep{raffel2020exploring}.
We follow the configuration used in T5-v1.1-base\footnote{\url{https://huggingface.co/docs/transformers/model_doc/t5v1.1}}. 
The vocabulary size of \method{} is $35,073$,  differing from the default configuration as we incorporate separate vocabulary for molecule SELFIES and protein amino acids.
In total, the \method{} model comprises $252$M parameters.

\begin{table*}[t!]
\centering
% \vspace{-1.3cm}
\resizebox{0.9\textwidth}{!}{
\tiny
\begin{tabular}{lcccccccccccccc}
\toprule
Dataset & BBBP & Tox21 & ClinTox & HIV & BACE & SIDER & Avg \\
\textbf{\#Molecules} & 2039 & 7831 & 1478 & 41127 & 1513 & 1427 & -\\
\textbf{\#Tasks} & 1 & 12 & 2 & 1 & 1 & 27 & -\\
\midrule
G-Contextual& 70.3$\pm$1.6 & 75.2$\pm$0.3 & 59.9$\pm$8.2 & 75.9$\pm$0.9 & 79.2$\pm$0.3 & 58.4$\pm$0.6 & 69.8 \\
G-Motif & 66.4$\pm$3.4 & 73.2$\pm$0.8 & 77.8$\pm$2.0 & 73.8$\pm$1.4 & 73.4$\pm$4.0 & 60.6$\pm$1.1 & 70.9\\
GROVER$_{\rm base}$ & 70.0$\pm$0.1 & 74.3$\pm$0.1 & 81.2$\pm$3.0 & 62.5$\pm$0.9 & 82.6$\pm$0.7 & 64.8$\pm$0.6 & 72.6 \\
GROVER$_{\rm large}$ & 69.5$\pm$0.1 & 73.5$\pm$0.1 & 76.2$\pm$3.7 & 68.2$\pm$1.1 & 81.0$\pm$1.4 & 65.4$\pm$0.1 & 72.3\\
GraphMVP & 72.4$\pm$1.6 & 75.9$\pm$0.5 & 79.1$\pm$2.8 & 77.0$\pm$1.2 & 81.2$\pm$0.9 & 63.9$\pm$1.2 & 74.9\\
MGSSL& 70.5$\pm$1.1 & 76.5$\pm$0.3 & 80.7$\pm$2.1 & 79.5$\pm$1.1 & 79.7$\pm$0.8 & 61.8$\pm$0.8 & 74.8\\
MolCLR & 72.2$\pm$2.1 & 75.0$\pm$0.2 & 91.2$\pm$3.5 & 78.1$\pm$0.5 & 82.4$\pm$0.9 & 58.9$\pm$1.4 & 76.3\\
GEM & 72.4$\pm$0.4 & \textbf{78.1$\pm$0.1} & 90.1$\pm$1.3 & \underline{80.6 $\pm$ 0.9} & 85.6$\pm$1.1 & 67.2$\pm$0.4 & 79.0 \\
\midrule
KV-PLM &  74.6$\pm$0.9 & 72.7$\pm$0.6 & -- & 74.0$\pm$1.2 & -- & 61.5$\pm$1.5 & --\\
Galactica& 66.1 & 68.9 & 82.6 & 74.5 & 61.7 & 63.2 & 69.5 \\
MoMu & 70.5$\pm$2.0 & 75.6$\pm$0.3 & 79.9$\pm$4.1 & 76.2$\pm$0.9  & 77.1$\pm$1.4 & 60.5$\pm$0.9 & 73.3 \\
MolXPT & \textbf{80.0 $\pm$ 0.5} &  77.1$\pm$0.2  & \underline{95.3 $\pm$ 0.2} & 78.1$\pm$0.4 & \underline{88.4 $\pm$ 1.0} & \underline{71.7 $\pm$ 0.2} & \underline{81.9} \\
\midrule
\method & \underline{77.7$\pm$0.6} & \underline{77.9$\pm$0.2} & \textbf{95.4$\pm$0.5} & \textbf{81.0$\pm$0.1} & \textbf{89.4$\pm$0.3} & \textbf{73.2$\pm$0.2} & \textbf{82.4}\\
\bottomrule
\end{tabular}}
\caption{Performance comparison on MoleculeNet (\textbf{Best}, \underline{Second Best}). The evaluation metric is AUROC. The baseline results are mainly sourced from MolXPT~\citep{liu2023molxpt}.}
\label{tab:moleculenet}
% \vspace{-0.6cm}
\end{table*}

\begin{table}[h!]
\centering
\resizebox{\linewidth}{!}{
\small
\begin{tabular}{cccc}
\toprule
Model & \#Params. & Solubility & Localization \\
\midrule
DDE & 205.3K & 59.77 $\pm$ 1.21 & 77.43 $\pm$ 0.42 \\
Moran & 123.4K & 57.73 $\pm$ 1.33 & 55.63 $\pm$ 0.85 \\
\midrule
LSTM & 26.7M & 70.18 $\pm$ 0.63 & 88.11 $\pm$ 0.14 \\
Transformer & 21.3M & 70.12 $\pm$ 0.31 & 75.74 $\pm$ 0.74 \\
CNN & 5.4M & 64.43 $\pm$ 0.25 & 82.67 $\pm$ 0.32 \\
ResNet & 11.0M & 67.33 $\pm$ 1.46 & 78.99 $\pm$ 4.41 \\
\midrule
ProtBert & 419.9M & 68.15 $\pm$ 0.92 & 91.32 $\pm$ 0.89 \\
ProtBert* & 419.9M & 59.17 $\pm$ 0.21 & 81.54 $\pm$ 0.09 \\
ESM-1b & 652.4M & \underline{70.23 $\pm$ 0.75} & \textbf{92.40 $\pm$ 0.35} \\
ESM-1b* & 652.4M & 67.02 $\pm$ 0.40 & 91.61 $\pm$ 0.10\\
\midrule
\method & 252.1M & \textbf{74.65 $\pm$ 0.49} & \underline{91.69 $\pm$ 0.05} \\
\bottomrule
\end{tabular}
}
\caption{Performance comparison of different methods on solubility and localization prediction tasks (\textbf{Best}, \underline{Second Best}). The evaluation metric is accuracy. * represents only tuning the prediction head. The baseline results are sourced from PEER~\citep{xu2022peer}.}
\label{tab:protein_property}
% \vspace{-0.5cm}
\end{table}

\begin{table*}[t!]
\centering
% \vspace{-1.3cm}
\resizebox{\textwidth}{!}{
\begin{tabular}{c  ccc  cc  ccc}
\toprule
& \multicolumn{3}{c}{BioSNAP} & \multicolumn{2}{c}{Human} & \multicolumn{3}{c}{BindingDB} \\
\cmidrule(r){2-4}  \cmidrule(r){5-6}  \cmidrule(r){7-9} Method & 
AUROC & AUPRC & Accuracy & AUROC & AUPRC & AUROC & AUPRC & Accuracy\\ \midrule

SVM             & 0.862$\pm$0.007       & 0.864$\pm$0.004  & 0.777$\pm$0.011  & 0.940$\pm$0.006 & 0.920$\pm$0.009 & 0.939$\pm$0.001 & 0.928$\pm$0.002 & 0.825$\pm$0.004\\
RF        & 0.860$\pm$0.005       & 0.886$\pm$0.005  & 0.804$\pm$0.005 & 0.952$\pm$0.011 & 0.953$\pm$0.010  & 0.942$\pm$0.011        & 0.921$\pm$0.016   & 0.880$\pm$0.012\\
DeepConv-DTI           & 0.886$\pm$0.006 & 0.890$\pm$0.006  & 0.805$\pm$0.009  & 0.980$\pm$0.002 & 0.981$\pm$0.002  & 0.945$\pm$0.002       & 0.925$\pm$0.005 &0.882$\pm$0.007\\
GraphDTA      & 0.887$\pm$0.008       & 0.890$\pm$0.007  & 0.800$\pm$0.007 & 0.981$\pm$0.001 & \underline{0.982$\pm$0.002}  & 0.951$\pm$0.002        & 0.934$\pm$0.002 & 0.888$\pm$0.005\\
MolTrans & 0.895$\pm$0.004       & 0.897$\pm$0.005  & 0.825$\pm$0.010 & 0.980$\pm$0.002 & 0.978$\pm$0.003  & 0.952$\pm$0.002        & 0.936$\pm$0.001 & 0.887$\pm$0.006 \\
DrugBAN & \underline{0.903$\pm$0.005}       & \underline{0.902$\pm$0.004}  & \underline{0.834$\pm$0.008} & \underline{0.982$\pm$0.002} & 0.980$\pm$0.003 & \underline{0.960$\pm$0.001}        & \underline{0.948$\pm$0.002} & \underline{0.904$\pm$0.004}\\ \midrule
\method & \textbf{0.937$\pm$0.001} & \textbf{0.937$\pm$0.004} & \textbf{0.874$\pm$0.001} & \textbf{0.989$\pm$0.001} & \textbf{0.985$\pm$0.002} & \textbf{0.963$\pm$0.001} & \textbf{0.952$\pm$0.001} & \textbf{0.907$\pm$0.003}\\
\bottomrule
\end{tabular}
}
\caption{Performance comparison on the BindingDB, Human and BioSNAP datasets. (\textbf{Best}, \underline{Second Best}). The baseline results derive from DrugBAN~\citep{bai2023interpretable}.}
\label{tab:dti}
% \vspace{-0.5cm}
\end{table*}

\begin{table}[t]
\centering
% \vspace{-0.5cm}
\resizebox{\linewidth}{!}{
\small
\begin{tabular}{cccc}
\toprule
Model & \#Params. & Yeast & Human \\
\midrule
DDE & 205.3K & 55.83 $\pm$ 3.13 & 62.77 $\pm$ 2.30 \\
Moran & 123.4K & 53.00 $\pm$ 0.50 & 54.67 $\pm$ 4.43 \\
\midrule
LSTM & 26.7M & 53.62 $\pm$ 2.72 & 63.75 $\pm$ 5.12 \\
Transformer & 21.3M & 54.12 $\pm$ 1.27 & 59.58 $\pm$ 2.09 \\
CNN & 5.4M & 55.07 $\pm$ 0.02 & 62.60 $\pm$ 1.67 \\
ResNet & 11.0M & 48.91 $\pm$ 1.78 & 68.61 $\pm$ 3.78 \\
\midrule
ProtBert & 419.9M & 63.72 $\pm$ 2.80 & 77.32 $\pm$ 1.10 \\
ProtBert* & 419.9M & 53.87 $\pm$ 0.38 & 83.61 $\pm$ 1.34 \\
ESM-1b & 652.4M & 57.00 $\pm$ 6.38 & 78.17 $\pm$ 2.91 \\
ESM-1b* & 652.4M & \textbf{66.07 $\pm$ 0.58} & \textbf{88.06 $\pm$ 0.24}\\
\midrule
\method & 252.1M & \underline{64.89 $\pm$ 0.43} & \underline{86.22 $\pm$ 0.53} \\
\bottomrule
\end{tabular}
}
\caption{Performance comparison on Yeast and Human datasets (\textbf{Best}, \underline{Second Best}). The evaluation metric is accuracy. * represents only tuning the prediction head. The baseline results derive from PEER~\citep{xu2022peer}.}
\label{tab:ppi}
% \vspace{-0.6cm}
\end{table}

\noindent{\textbf{Pre-training}}
During the pre-training phase, the model is trained in a multi-task way on six tasks that can be classified into three types:
(1) Applying T5 objective to each single modality including molecule SELFIES (task \#1), protein FASTA (task \#2), and general text (task \#3) independently.
(2) Applying T5 objective to wrapped text from scientific corpus (task \#4).
(3) Bidirectional translation for the molecule SELFIES-text pairs (task \#5) and protein FASTA-text pairs (task \#6).
By effectively learning the underlying connections and properties of bio-entities from textual information through these pre-training tasks, \method{} allows for a holistic understanding of the biological domain, thereby facilitating enhanced prediction and generation abilities in various biological tasks.

\noindent{\textbf{Fine-tuning}}
\label{sec:finetune}
\method{} can be fine-tuned on various downstream tasks involving molecules, proteins, and text.
To unify different downstream tasks and reduce the gap between pre-training and fine-tuning~\citep{brown2020language} stage, we adopt the prompt-based fine-tuning~\citep{DBLP:conf/acl/GaoFC20} approach, which facilitates various task formats into a sequence generation format.

\section{Experiments and Results}
We evaluate \method{} on $15$ well-established downstream tasks, which can be categorized into three types: single-instance prediction, multi-instance prediction, and cross-modal generation.
We include details regarding fine-tuning datasets, baselines, and prompts in Appendix~\ref{sec:finetune_detail}.

For the downstream binary classification tasks presented in Section~\ref{sec:single_instance} and ~\ref{sec:multi_instance}, the calculation of evaluation metrics such as AUROC and AUPRC necessitates the soft probability of the predicted label.
As we use the prompt-based fine-tuning method, the output is either \textit{Yes} for the positive label or \textit{No} for the negative label.
To obtain an appropriate label distribution, following~\citet{liu2023molxpt}, we first extract the probabilities of \textit{Yes} and \textit{No} tokens (denoted as $p_{pos}$ and $p_{neg}$ respectively) and normalize them.
The resulting probability for positive label is $\frac{p_{pos}}{p_{pos}+p_{neg}}$ and negative label is $\frac{p_{neg}}{p_{pos}+p_{neg}}$.

\subsection{Single-instance Prediction}
\label{sec:single_instance}
\subsubsection{Molecule Property Prediction}
\label{sec:molecule_property}
Molecule property prediction aims to determine whether a given molecule exhibits specific properties.
MoleculeNet~\citep{wu2018moleculenet} is a widely used benchmark for molecule property prediction, encompassing diverse datasets that cover numerous molecular aspects, such as quantum mechanics, physical chemistry, biophysics, etc.
In line with~\citet{liu2023molxpt}, we conduct experiments on six binary classification tasks, including BBBP, Tox21, ClinTox, HIV, BACE, and SIDER.
Following~\citep{fang2022geometry}, we adopt the scaffold splitting, which is more challenging compared to random splitting.

\noindent{\textbf{Baselines}}
We compare \method{} with two types of baselines:
(1) pre-trained Graph Neural Network (GNN) using molecular graph as input, which are G-Contextual~\citep{rong2020self}, G-Motif~\citep{rong2020self}, GROVER$_{\rm base}$~\citep{rong2020self}, GROVER$_{\rm large}$~\citep{rong2020self}, GraphMVP~\citep{DBLP:conf/iclr/LiuWLLGT22}, MGSSL~\citep{zhang2021motif} MolCLR~\citep{wang2022molecular} and GEM~\citep{fang2022geometry};
(2) pre-trained language model baselines, which are KV-PLM~\citep{zeng2022deep}, Galactica~\citep{taylor2022galactica}, MoMu~\citep{su2022molecular} and MolXPT~\cite{liu2023molxpt}.

\noindent{\textbf{Results}}
The results are presented in Table~\ref{tab:moleculenet} with all statistics derived from three random runs.
From these results, we can see that \method{} surpasses baselines on most downstream tasks in MoleculeNet.
\method{} exhibits superior performance compared to GNN baselines that are pre-trained on 2D/3D molecular data, underscoring the effectiveness of knowledge in text.
Furthermore, \method{} outperforms other language model baselines, which may be attributed to the presence of molecule property descriptions in scientific contextual text or existing biological database entries.

\subsubsection{Protein Property Prediction}
\label{sec:protein_property_prediction}
Protein property prediction is crucial as it provides critical insights into the behavior and functions of proteins.
We concentrate on two protein property prediction tasks on PEER benchmark~\citep{xu2022peer}: protein solubility prediction, which aims to predict whether the given protein is soluble, and protein localization prediction, which is to classify proteins as either ``membrane-bound'' or ``soluble''.

\noindent{\textbf{Baselines}}
We compare \method{} with three types of baselines provided in PEER benchmark:
(1) feature engineers, including two protein sequence feature descriptors: Dipeptide Deviation from Expected Mean (DDE)~\citep{saravanan2015harnessing} and Moran correlation (Moran)~\citep{feng2000prediction};
(2) protein sequence encoders, including LSTM~\citep{hochreiter1997long}, Transformers~\citep{vaswani2017attention}, CNN~\citep{o2015introduction} and ResNet~\citep{he2016deep};
(3) pre-trained protein language models, which are pre-trained using extensive collections of protein FASTA sequences, including ProtBert~\citep{elnaggar2021prottrans} and ESM-1b~\citep{rives2021biological}.
Both ProtBert and ESM-1b are studied with two settings 
(i) freezing the protein language model parameters and only training the prediction head;
(ii) fine-tuning all model parameters.

\noindent{\textbf{Results}}
The results are displayed in Table~\ref{tab:protein_property}, with all statistics derived from three random runs.
In the protein solubility prediction task, \method{} outperforms all baselines in PEER~\citep{xu2022peer} benchmark.
In the protein localization prediction task, \method{} is the second best among all methods.
Notably, ProtBert and ESM-1b are both pre-trained on a large corpus of protein sequences, which is comparable to or even larger than ours.
Moreover, these models are two to three times larger than \method{}.
These demonstrate the potential of \method{} for enhanced predictive capabilities in protein property prediction by integrating textual information.

\subsection{Multi-instance Prediction}
\label{sec:multi_instance}
\subsubsection{Drug-target Interaction Prediction}
\label{sec:dti}
Drug-target interaction (DTI) prediction plays a crucial role in drug discovery, as it aims to predict whether a given drug (molecule) and target (protein) can interact with each other.
We select three widely-used DTI datasets with a binary classification setting, which are BioSNAP~\citep{zitnik2018biosnap}, BindingDB~\citep{liu2007bindingdb} and Human~\citep{liu2015improving,chen2020transformercpi}.

\noindent{\textbf{Baselines}}
We compare \method{} with two types of baselines:
(1) traditional machine learning methods including SVM~\citep{cortes1995support} and Random Forest (RF)~\citep{ho1995random};
(2) deep learning methods including DeepConv-DTI~\citep{Lee2019DeepConvDTIPO}, GraphDTA~\citep{Nguyen2020GraphDTAPD}, MolTrans~\citep{Huang2021MolTransMI} and DrugBAN~\citep{bai2023interpretable}, in which drug and target feature are firstly extracted by well-design drug encoder and protein encoder then fused for prediction.

\noindent{\textbf{Results}}
The results on BioSNAP, Human, and BindingDB datasets are presented in Table~\ref{tab:dti}. All statistics are obtained from five random runs.
On BioSNAP and BindingDB datasets, \method{} consistently outperforms other methods in various performance metrics, including AUROC, AUPRC, and accuracy.
For the Human dataset, although deep learning-based models generally exhibit strong performance, the \method{} model demonstrates a slight advantage over the baseline models.
It is worth noting that, in contrast to most deep learning-based baselines, our \method{} does not rely on a specific design tailored for molecules or proteins. A possible explanation for the superior performance of \method{} is that the SELFIES and FASTA representations effectively capture the intricate structure and function of molecules and proteins, and the interaction information between them may be well-described in the contextual scientific literature or corresponding text entries in databases.

\begin{table*}[t]
% \vspace{-1.2cm}
\resizebox{\textwidth}{!}{
\centering
\small
\begin{tabular}{ccccccccc}
\toprule
Model & \#Params. & BLEU-2 & BLEU-4 & ROUGE-1 & ROUGE-2 & ROUGE-L & METEOR & Text2Mol \\
\midrule
% Ground Truth & - & - & - & - & - & - & - & 0.609 \\\midrule
RNN & 56M & 0.251 & 0.176 & 0.450 & 0.278 & 0.394 & 0.363 & 0.426 \\
Transformer & 76M & 0.061 & 0.027 & 0.204 & 0.087 & 0.186 & 0.114 & 0.057 \\\midrule
T5-small & 77M & 0.501 & 0.415 & 0.602 & 0.446 & 0.545 & 0.532 & 0.526 \\
T5-base & 248M & 0.511 & 0.423 & 0.607 & 0.451 & 0.550 & 0.539 & 0.523 \\
T5-large & 783M & 0.558 & 0.467 & 0.630 & 0.478 & 0.569 & 0.586 & 0.563 \\
\midrule
% T5-small & 77M & 0.501 & 0.415 & 0.602 & 0.446 & 0.545 & 0.532 & 0.526 \\
MolT5-small & 77M & 0.519 & 0.436 & 0.620 & 0.469 & 0.563 & 0.551 & 0.540 \\
% T5-base & 248M & 0.511 & 0.423 & 0.607 & 0.451 & 0.550 & 0.539 & 0.523 \\
MolT5-base & 248M & 0.540 & 0.457 & 0.634 & 0.485 & 0.578 & 0.569 & 0.547 \\
% T5-large & 783M & 0.558 & 0.467 & 0.630 & 0.478 & 0.569 & 0.586 & 0.563 \\
MolT5-large & 783M & \underline{0.594} & \underline{0.508} & 0.654 & 0.510 & 0.594 & 0.614 & 0.582 \\\midrule
GPT-3.5-turbo (zero-shot) & >175B & 0.103 & 0.050 & 0.261 & 0.088 & 0.204 & 0.161 & 0.352\\
GPT-3.5-turbo (10-shot MolReGPT) & >175B & 0.565 & 0.482 & 0.623 & 0.450 & 0.543 & 0.585 & 0.560\\\midrule
MolXPT & 350M & \underline{0.594} & 0.505 & \underline{0.660} & \underline{0.511} & \underline{0.597} & \underline{0.626} & \underline{0.594} \\\midrule
\method & 252M & \textbf{0.635} & \textbf{0.556} & \textbf{0.692} & \textbf{0.559} & \textbf{0.633} & \textbf{0.656} & \textbf{0.603} \\
\bottomrule
\end{tabular}
}
\caption{Performance comparison on molecule captioning task (\textbf{Best}, \underline{Second Best}). Rouge scores are F1 values. 
The Text2Mol score between ground truth molecule and corresponding text description is $0.609$.
The baseline results derive from MolT5~\citep{DBLP:conf/emnlp/EdwardsLRHCJ22}, MolXPT~\citep{liu2023molxpt}, and MolReGPT~\citep{li2023empowering}.
}
\label{tab:mol2text}
% \vspace{-0.3cm}
\end{table*}

\begin{table*}[t]
\resizebox{\textwidth}{!}{
\centering
\begin{tabular}{ccccccccccc}
\toprule
Model & \#Params. & BLEU$\uparrow$ & Exact$\uparrow$ & Levenshtein$\downarrow$ & MACCS FTS$\uparrow$ & RDK FTS$\uparrow$ & Morgan FTS$\uparrow$ & FCD$\downarrow$ & Text2Mol$\uparrow$ & Validity$\uparrow$ \\
\midrule
% Ground Truth & - & 1.000 & 1.000 & 0.00 & 1.000 & 1.000 & 1.000 & 0.00 & 0.609 & 1.0 \\\midrule
RNN & 56M & 0.652 & 0.005 & 38.09 & 0.591 & 0.400 & 0.362 & 4.55 & 0.409 & 0.542 \\
Transformer & 76M & 0.499 & 0.000 & 57.66 & 0.480 & 0.320 & 0.217 & 11.32 & 0.277 & 0.906 \\\midrule
T5-small & 77M & 0.741 & 0.064 & 27.703 & 0.704 & 0.578  & 0.525  & 2.89  & 0.479 & 0.608 \\
T5-base & 248M & 0.762 & 0.069 & 24.950 & 0.731 &  0.605 & 0.545 & 2.48 & 0.499 & 0.660  \\
T5-large & 783M & 0.854 & 0.279 & 16.721 & 0.823 &  0.731 & 0.670 & 1.22 & 0.552 & 0.902 \\
\midrule
% T5-small & 77M & 0.741 & 0.064 & 27.703 & 0.704 & 0.578  & 0.525  & 2.89  & 0.479 & 0.608 \\
MolT5-small & 77M & 0.755 & 0.079 & 25.988 & 0.703 & 0.568 & 0.517 & 2.49 & 0.482 & 0.721 \\
% T5-base & 248M & 0.762 & 0.069 & 24.950 & 0.731 &  0.605 & 0.545 & 2.48 & 0.499 & 0.660  \\
MolT5-base & 248M & 0.769 & 0.081 & 24.458 & 0.721 &  0.588 & 0.529 & 2.18 & 0.496 & 0.772 \\
% T5-large & 783M & \underline{0.854} & 0.279 & 16.721 & 0.823 &  0.731 & 0.670 & 1.22 & 0.552 & 0.902 \\
MolT5-large & 783M & \underline{0.854} & \underline{0.311} & \underline{16.071} & 0.834 & 0.746 & \underline{0.684} & 1.20 & 0.554 & 0.905 \\\midrule
GPT-3.5-turbo (zero-shot) & >175B & 0.489 & 0.019 & 52.13 & 0.705 & 0.462 & 0.367 & 2.05 & 0.479 & 0.802 \\
GPT-3.5-turbo (10-shot MolReGPT) & >175B & 0.790 & 0.139 & 24.91 & 0.847 & 0.708 & 0.624 & 0.57 & 0.571 & 0.887\\\midrule
MolXPT & 350M & - & 0.215 & - & \underline{0.859} & \underline{0.757} & 0.667 & \underline{0.45} & \textbf{0.578} & \underline{0.983} \\\midrule
\method & 252M & \textbf{0.867} & \textbf{0.413} & \textbf{15.097} & \textbf{0.886} & \textbf{0.801} & \textbf{0.734} & \textbf{0.43} & \underline{0.576} & \textbf{1.000} \\
\bottomrule
\end{tabular}
}
\caption{Performance comparison on text-based molecule generation task (\textbf{Best}, \underline{Second Best}). Following~\citet{DBLP:conf/emnlp/EdwardsLRHCJ22}, BLEU, Exact, Levenshtein, and Validity are computed on all generated molecues while other metrics are computed only on syntactically valid molecules. 
The Text2Mol score for ground truth is $0.609$.
The baseline results derive from MolT5~\citep{DBLP:conf/emnlp/EdwardsLRHCJ22}, MolXPT~\citep{liu2023molxpt}, and MolReGPT~\citep{li2023empowering}.
}
\label{tab:text2mol}
% \vspace{-0.5cm}
\end{table*}

\subsubsection{Protein-protein Interaction Prediction}
\label{sec:ppi}
Protein-protein interaction (PPI) prediction plays a vital role in understanding protein functions and structures, as it aims to determine the potential interactions between pairs of proteins.
Following PEER~\citep{xu2022peer} benchmark, we perform fine-tuning on two PPI datasets: Yeast~\citep{guo2008using} and Human~\citep{pan2010large}.

\noindent{\textbf{Baselines}}
The baselines for comparison are the same as that in Section~\ref{sec:protein_property_prediction}.

\noindent{\textbf{Results}}
The results are shown in Table~\ref{tab:ppi}. All statistics are over three random runs.
On two PPI datasets, \method{} shows superior performance compared to almost all baseline models.
Remarkably, \method{} outperforms both ProtBert and ESM-1b (with full parameters fine-tuned).
This result strongly highlights the crucial role of incorporating textual information during the pre-training of BioT5, which effectively establishes profound connections between proteins.
Our model, despite being smaller, is able to harness the unstructured information embedded in scientific text and structured information from biological databases, encapsulating the comprehensive knowledge of proteins in their varying contexts. 

\subsection{Cross-modal Generation}
In this section, we evaluate the performance of \method{} on the cross-modal generation task.
Specifically, we fine-tune \method{} on molecule captioning and text-based molecule generation tasks.
These two tasks are proposed by MolT5~\citep{DBLP:conf/emnlp/EdwardsLRHCJ22} and both use the ChEBI-20 dataset~\citep{DBLP:conf/emnlp/EdwardsZJ21}.
The evaluation metrics and some interesting cases are introduced in Appendix~\ref{sec:mol_text_metric} and~\ref{sec:case_study}.
\subsubsection{Molecule Captioning}
\label{sec:mol2text}
For the given molecule, the goal of molecule captioning task is to provide a description of the given molecule.
As we use SELFIES sequences to represent molecules, this task can be formulated as an exotic sequence-to-sequence translation task.

\noindent{\textbf{Baselines}}
The baselines include:
RNN~\citep{medsker2001recurrent}, Transformer~\citep{vaswani2017attention}, T5~\citep{raffel2020exploring}, MolT5~\citep{DBLP:conf/emnlp/EdwardsLRHCJ22}, GPT-3.5-turbo\footnote{\url{https://openai.com/blog/openai-api}} with zero-shot and 10-shot MolReGPT~\citep{li2023empowering} settings, and MolXPT~\cite{liu2023molxpt}.

\noindent{\textbf{Results}}
The results are shown in Table~\ref{tab:mol2text}.
\method{} only has nearly the same number of parameters as MolT5-base, but outperforms all baseline models in all metrics, including those that have more parameters.
The Text2Mol score is $0.603$, which is very close to the Text2Mol score of $0.609$ between the ground truth molecule and the corresponding description.
We can attribute this superior performance to the unstructured contextual knowledge and structured database knowledge induced in \method{} pre-training, which helps the model learn the intricate relationship between text and molecules. 

\subsubsection{Text-Based Molecule Generation}
\label{sec:text2mol}
This is a reverse task of molecule captioning.
Given the nature language description of the intended molecule, the goal is to generate the molecule that fits the description.

\noindent{\textbf{Baselines}}
The compared baselines are the same as baselines in Section~\ref{sec:mol2text}.

\noindent{\textbf{Results}}
The results are presented in Table~\ref{tab:text2mol}.
\method{} only uses parameters similar to MolT5-base yet delivers superior performance across nearly all metrics.
Notably, the exact match score of \method{} surpasses the MolT5-Large by 32.8\% while maintaining a validity of $1.0$.
This indicates that \method{} not only generates more relevant molecules corresponding to the given text descriptions, but also ensures a 100\% validity for the generated molecules.
The overall enhanced performance of \method{} can be attributed to the incorporation of both contextual and database knowledge, as well as the utilization of SELFIES for molecular representation.

\section{Conclusions and Future Work}
In this paper, we propose \method{}, a comprehensive pre-training framework capable of capturing the underlying relations and properties of bio-entities by leveraging both structured and unstructured data sources with 100\% robust molecular representation. 
Our method effectively enriches cross-modal integration in biology with chemical knowledge and natural language associations, demonstrating notable improvements in various tasks.

For future work, we aim to further enrich our model by incorporating additional biological data types, such as genomics or transcriptomics data, to create a more holistic biological pre-training framework. 
Additionally, we plan to evaluate the interpretability of \method{} predictions, aiming to provide more insights into the biological systems under study. 
Thus, we foresee our work sparking further innovation in the use of AI models in the field of computational biology, ultimately leading to a deeper understanding of biological systems and facilitating more efficient drug discovery.

\section{Limitations}
One limitation of \method{} is conducting full-parameter fine-tuning on each downstream task. 
This is done because we do not observe generalization ability among different downstream tasks using instruction-tuning~\citep{DBLP:conf/iclr/WeiBZGYLDDL22} method.
Another reason is that combining data from different tasks using instructions results in data leakage. 
For example, have noticed overlaps between the training set of BindingDB and the test sets of BioSNAP and Human.
Additionally, we only demonstrate the ability of \method{} in text, molecule, and protein modalities. 
Numerous other biological modalities, such as DNA/RNA sequences and cells, exist, and there are many other tasks within a single modality or across multiple modalities.
Moreover, \method{} primarily focuses on the sequence format of bio-entities, yet other formats, such as 2D or 3D structures, also hold significant importance.
We leave further exploration of these to future work.

% additional
\section{Risks}
Despite the potential benefits of BioT5 in research and pharmaceutical applications, the risk of misuse should be prevented. 
The BioT5 might fail to generate effective molecules for treating specific diseases and could potentially produce compounds with adverse side effects. 
Additionally, the BioT5 could be employed to create dangerous molecules.

\section{Acknowledgements}
This work was supported by the National Key Research and Development Program of China (No. 2020YFB1406702), National Natural Science Foundation of China (NSFC Grant No. 62122089), Beijing Outstanding Young Scientist Program NO. BJJWZYJH012019100020098, and Intelligent Social Governance Platform, Major Innovation \& Planning Interdisciplinary Platform for the ``Double-First Class'' Initiative, Renmin University of China, the Fundamental Research Funds for the Central Universities, and the Research Funds of Renmin University of China. 

% \clearpage
\bibliography{custom}

\begin{thebibliography}{99}
\expandafter\ifx\csname natexlab\endcsname\relax\def\natexlab#1{#1}\fi

\bibitem[{uni(2023)}]{uniprot2023uniprot}
 2023.
\newblock Uniprot: the universal protein knowledgebase in 2023.
\newblock \emph{Nucleic Acids Research}, 51(D1):D523--D531.

\bibitem[{Armenteros et~al.(2017)Armenteros, S{\o}nderby, S{\o}nderby, Nielsen,
  and Winther}]{DBLP:journals/bioinformatics/ArmenterosSSNW17}
Jos{\'{e}} Juan~Almagro Armenteros, Casper~Kaae S{\o}nderby, S{\o}ren~Kaae
  S{\o}nderby, Henrik Nielsen, and Ole Winther. 2017.
\newblock \href {https://doi.org/10.1093/bioinformatics/btx431} {Deeploc:
  prediction of protein subcellular localization using deep learning}.
\newblock \emph{Bioinform.}, 33(21):3387--3395.

\bibitem[{AstraZeneca(2023)}]{future_smallM}
AstraZeneca. 2023.
\newblock \href
  {https://www.astrazeneca.com/r-d/next-generation-therapeutics/small-molecule.html}
  {A big future for small molecules: targeting the undruggable}.

\bibitem[{Bagal et~al.(2022)Bagal, Aggarwal, Vinod, and
  Priyakumar}]{DBLP:journals/jcisd/BagalAVP22}
Viraj Bagal, Rishal Aggarwal, P.~K. Vinod, and U.~Deva Priyakumar. 2022.
\newblock \href {https://doi.org/10.1021/acs.jcim.1c00600} {Molgpt: Molecular
  generation using a transformer-decoder model}.
\newblock \emph{J. Chem. Inf. Model.}, 62(9):2064--2076.

\bibitem[{Bai et~al.(2021)Bai, Miljkovi{\'c}, Ge, Greene, John, and
  Lu}]{bai2021hierarchical}
Peizhen Bai, Filip Miljkovi{\'c}, Yan Ge, Nigel Greene, Bino John, and Haiping
  Lu. 2021.
\newblock Hierarchical clustering split for low-bias evaluation of drug-target
  interaction prediction.
\newblock In \emph{2021 IEEE International Conference on Bioinformatics and
  Biomedicine (BIBM)}, pages 641--644. IEEE.

\bibitem[{Bai et~al.(2023)Bai, Miljkovi{\'c}, John, and
  Lu}]{bai2023interpretable}
Peizhen Bai, Filip Miljkovi{\'c}, Bino John, and Haiping Lu. 2023.
\newblock Interpretable bilinear attention network with domain adaptation
  improves drug--target prediction.
\newblock \emph{Nature Machine Intelligence}, 5(2):126--136.

\bibitem[{Banerjee and Lavie(2005)}]{DBLP:conf/acl/BanerjeeL05}
Satanjeev Banerjee and Alon Lavie. 2005.
\newblock \href {https://aclanthology.org/W05-0909/} {{METEOR:} an automatic
  metric for {MT} evaluation with improved correlation with human judgments}.
\newblock In \emph{Proceedings of the Workshop on Intrinsic and Extrinsic
  Evaluation Measures for Machine Translation and/or Summarization@ACL 2005,
  Ann Arbor, Michigan, USA, June 29, 2005}, pages 65--72. Association for
  Computational Linguistics.

\bibitem[{Beltagy et~al.(2019)Beltagy, Lo, and
  Cohan}]{DBLP:conf/emnlp/BeltagyLC19}
Iz~Beltagy, Kyle Lo, and Arman Cohan. 2019.
\newblock \href {https://doi.org/10.18653/v1/D19-1371} {Scibert: {A} pretrained
  language model for scientific text}.
\newblock In \emph{Proceedings of the 2019 Conference on Empirical Methods in
  Natural Language Processing and the 9th International Joint Conference on
  Natural Language Processing, {EMNLP-IJCNLP} 2019, Hong Kong, China, November
  3-7, 2019}, pages 3613--3618. Association for Computational Linguistics.

\bibitem[{Boutet et~al.(2007)Boutet, Lieberherr, Tognolli, Schneider, and
  Bairoch}]{boutet2007uniprotkb}
Emmanuel Boutet, Damien Lieberherr, Michael Tognolli, Michel Schneider, and
  Amos Bairoch. 2007.
\newblock Uniprotkb/swiss-prot: the manually annotated section of the uniprot
  knowledgebase.
\newblock \emph{Plant bioinformatics: methods and protocols}, pages 89--112.

\bibitem[{Brister et~al.(2015)Brister, Ako-Adjei, Bao, and
  Blinkova}]{brister2015ncbi}
J~Rodney Brister, Danso Ako-Adjei, Yiming Bao, and Olga Blinkova. 2015.
\newblock Ncbi viral genomes resource.
\newblock \emph{Nucleic acids research}, 43(D1):D571--D577.

\bibitem[{Brown et~al.(2020)Brown, Mann, Ryder, Subbiah, Kaplan, Dhariwal,
  Neelakantan, Shyam, Sastry, Askell et~al.}]{brown2020language}
Tom Brown, Benjamin Mann, Nick Ryder, Melanie Subbiah, Jared~D Kaplan, Prafulla
  Dhariwal, Arvind Neelakantan, Pranav Shyam, Girish Sastry, Amanda Askell,
  et~al. 2020.
\newblock Language models are few-shot learners.
\newblock \emph{Advances in neural information processing systems},
  33:1877--1901.

\bibitem[{Butina(1999)}]{DBLP:journals/jcisd/Butina99}
Darko Butina. 1999.
\newblock \href {https://doi.org/10.1021/ci9803381} {Unsupervised data base
  clustering based on daylight's fingerprint and tanimoto similarity: {A} fast
  and automated way to cluster small and large data sets}.
\newblock \emph{J. Chem. Inf. Comput. Sci.}, 39(4):747--750.

\bibitem[{Canese and Weis(2013)}]{canese2013pubmed}
Kathi Canese and Sarah Weis. 2013.
\newblock Pubmed: the bibliographic database.
\newblock \emph{The NCBI handbook}, 2(1).

\bibitem[{Cao et~al.(2013)Cao, Xu, and
  Liang}]{DBLP:journals/bioinformatics/CaoXL13}
Dong{-}Sheng Cao, Qing{-}Song Xu, and Yi{-}Zeng Liang. 2013.
\newblock \href {https://doi.org/10.1093/bioinformatics/btt072} {propy: a tool
  to generate various modes of chou's pseaac}.
\newblock \emph{Bioinform.}, 29(7):960--962.

\bibitem[{Chen et~al.(2020)Chen, Tan, Wang, Zhong, Liu, Yang, Luo, Chen, Jiang,
  and Zheng}]{chen2020transformercpi}
Lifan Chen, Xiaoqin Tan, Dingyan Wang, Feisheng Zhong, Xiaohong Liu, Tianbiao
  Yang, Xiaomin Luo, Kaixian Chen, Hualiang Jiang, and Mingyue Zheng. 2020.
\newblock Transformercpi: improving compound--protein interaction prediction by
  sequence-based deep learning with self-attention mechanism and label reversal
  experiments.
\newblock \emph{Bioinformatics}, 36(16):4406--4414.

\bibitem[{Chithrananda et~al.(2020)Chithrananda, Grand, and
  Ramsundar}]{chithrananda2020chemberta}
Seyone Chithrananda, Gabriel Grand, and Bharath Ramsundar. 2020.
\newblock Chemberta: Large-scale self-supervised pretraining for molecular
  property prediction.
\newblock \emph{arXiv preprint arXiv:2010.09885}.

\bibitem[{Cortes and Vapnik(1995)}]{cortes1995support}
Corinna Cortes and Vladimir Vapnik. 1995.
\newblock Support-vector networks.
\newblock \emph{Machine learning}, 20(3):273--297.

\bibitem[{Dara et~al.(2022)Dara, Dhamercherla, Jadav, Babu, and
  Ahsan}]{dara2022machine}
Suresh Dara, Swetha Dhamercherla, Surender~Singh Jadav, CH~Madhu Babu, and
  Mohamed~Jawed Ahsan. 2022.
\newblock Machine learning in drug discovery: a review.
\newblock \emph{Artificial Intelligence Review}, 55(3):1947--1999.

\bibitem[{Durant et~al.(2002)Durant, Leland, Henry, and
  Nourse}]{durant2002reoptimization}
Joseph~L Durant, Burton~A Leland, Douglas~R Henry, and James~G Nourse. 2002.
\newblock Reoptimization of mdl keys for use in drug discovery.
\newblock \emph{Journal of chemical information and computer sciences},
  42(6):1273--1280.

\bibitem[{Edwards et~al.(2022)Edwards, Lai, Ros, Honke, Cho, and
  Ji}]{DBLP:conf/emnlp/EdwardsLRHCJ22}
Carl Edwards, Tuan~Manh Lai, Kevin Ros, Garrett Honke, Kyunghyun Cho, and Heng
  Ji. 2022.
\newblock \href {https://aclanthology.org/2022.emnlp-main.26} {Translation
  between molecules and natural language}.
\newblock In \emph{Proceedings of the 2022 Conference on Empirical Methods in
  Natural Language Processing, {EMNLP} 2022, Abu Dhabi, United Arab Emirates,
  December 7-11, 2022}, pages 375--413. Association for Computational
  Linguistics.

\bibitem[{Edwards et~al.(2021)Edwards, Zhai, and
  Ji}]{DBLP:conf/emnlp/EdwardsZJ21}
Carl Edwards, ChengXiang Zhai, and Heng Ji. 2021.
\newblock \href {https://doi.org/10.18653/v1/2021.emnlp-main.47} {Text2mol:
  Cross-modal molecule retrieval with natural language queries}.
\newblock In \emph{Proceedings of the 2021 Conference on Empirical Methods in
  Natural Language Processing, {EMNLP} 2021, Virtual Event / Punta Cana,
  Dominican Republic, 7-11 November, 2021}, pages 595--607. Association for
  Computational Linguistics.

\bibitem[{Elnaggar et~al.(2021)Elnaggar, Heinzinger, Dallago, Rehawi, Wang,
  Jones, Gibbs, Feher, Angerer, Steinegger et~al.}]{elnaggar2021prottrans}
Ahmed Elnaggar, Michael Heinzinger, Christian Dallago, Ghalia Rehawi, Yu~Wang,
  Llion Jones, Tom Gibbs, Tamas Feher, Christoph Angerer, Martin Steinegger,
  et~al. 2021.
\newblock Prottrans: Toward understanding the language of life through
  self-supervised learning.
\newblock \emph{IEEE transactions on pattern analysis and machine
  intelligence}, 44(10):7112--7127.

\bibitem[{Fang et~al.(2022)Fang, Liu, Lei, He, Zhang, Zhou, Wang, Wu, and
  Wang}]{fang2022geometry}
Xiaomin Fang, Lihang Liu, Jieqiong Lei, Donglong He, Shanzhuo Zhang, Jingbo
  Zhou, Fan Wang, Hua Wu, and Haifeng Wang. 2022.
\newblock Geometry-enhanced molecular representation learning for property
  prediction.
\newblock \emph{Nature Machine Intelligence}, 4(2):127--134.

\bibitem[{Feng and Zhang(2000)}]{feng2000prediction}
Zhi-Ping Feng and Chun-Ting Zhang. 2000.
\newblock Prediction of membrane protein types based on the hydrophobic index
  of amino acids.
\newblock \emph{Journal of protein chemistry}, 19:269--275.

\bibitem[{Ferruz et~al.(2022)Ferruz, Schmidt, and
  H{\"o}cker}]{ferruz2022protgpt2}
Noelia Ferruz, Steffen Schmidt, and Birte H{\"o}cker. 2022.
\newblock Protgpt2 is a deep unsupervised language model for protein design.
\newblock \emph{Nature communications}, 13(1):4348.

\bibitem[{Gao et~al.(2021)Gao, Fisch, and Chen}]{DBLP:conf/acl/GaoFC20}
Tianyu Gao, Adam Fisch, and Danqi Chen. 2021.
\newblock \href {https://doi.org/10.18653/v1/2021.acl-long.295} {Making
  pre-trained language models better few-shot learners}.
\newblock In \emph{Proceedings of the 59th Annual Meeting of the Association
  for Computational Linguistics and the 11th International Joint Conference on
  Natural Language Processing, {ACL/IJCNLP} 2021, (Volume 1: Long Papers),
  Virtual Event, August 1-6, 2021}, pages 3816--3830. Association for
  Computational Linguistics.

\bibitem[{Gu et~al.(2021)Gu, Tinn, Cheng, Lucas, Usuyama, Liu, Naumann, Gao,
  and Poon}]{gu2021domain}
Yu~Gu, Robert Tinn, Hao Cheng, Michael Lucas, Naoto Usuyama, Xiaodong Liu,
  Tristan Naumann, Jianfeng Gao, and Hoifung Poon. 2021.
\newblock Domain-specific language model pretraining for biomedical natural
  language processing.
\newblock \emph{ACM Transactions on Computing for Healthcare (HEALTH)},
  3(1):1--23.

\bibitem[{Guo et~al.(2008)Guo, Yu, Wen, and Li}]{guo2008using}
Yanzhi Guo, Lezheng Yu, Zhining Wen, and Menglong Li. 2008.
\newblock Using support vector machine combined with auto covariance to predict
  protein--protein interactions from protein sequences.
\newblock \emph{Nucleic acids research}, 36(9):3025--3030.

\bibitem[{Hastings et~al.(2016)Hastings, Owen, Dekker, Ennis, Kale,
  Muthukrishnan, Turner, Swainston, Mendes, and Steinbeck}]{hastings2016chebi}
Janna Hastings, Gareth Owen, Adriano Dekker, Marcus Ennis, Namrata Kale,
  Venkatesh Muthukrishnan, Steve Turner, Neil Swainston, Pedro Mendes, and
  Christoph Steinbeck. 2016.
\newblock Chebi in 2016: Improved services and an expanding collection of
  metabolites.
\newblock \emph{Nucleic acids research}, 44(D1):D1214--D1219.

\bibitem[{He et~al.(2016)He, Zhang, Ren, and Sun}]{he2016deep}
Kaiming He, Xiangyu Zhang, Shaoqing Ren, and Jian Sun. 2016.
\newblock Deep residual learning for image recognition.
\newblock In \emph{Proceedings of the IEEE conference on computer vision and
  pattern recognition}, pages 770--778.

\bibitem[{Heller et~al.(2013)Heller, McNaught, Stein, Tchekhovskoi, and
  Pletnev}]{heller2013inchi}
Stephen Heller, Alan McNaught, Stephen Stein, Dmitrii Tchekhovskoi, and Igor
  Pletnev. 2013.
\newblock Inchi-the worldwide chemical structure identifier standard.
\newblock \emph{Journal of cheminformatics}, 5(1):1--9.

\bibitem[{Ho(1995)}]{ho1995random}
Tin~Kam Ho. 1995.
\newblock Random decision forests.
\newblock In \emph{Proceedings of 3rd International Conference on Document
  Analysis and Recognition}, volume~1, pages 278--282.

\bibitem[{Hochreiter and Schmidhuber(1997)}]{hochreiter1997long}
Sepp Hochreiter and J{\"u}rgen Schmidhuber. 1997.
\newblock Long short-term memory.
\newblock \emph{Neural computation}, 9(8):1735--1780.

\bibitem[{Huang et~al.(2021)Huang, Xiao, Glass, and Sun}]{Huang2021MolTransMI}
Kexin Huang, Cao Xiao, Lucas Glass, and Jimeng Sun. 2021.
\newblock {MolTrans}: Molecular interaction transformer for drug–target
  interaction prediction.
\newblock \emph{Bioinformatics}, 37:830 -- 836.

\bibitem[{Irwin et~al.(2020)Irwin, Tang, Young, Dandarchuluun, Wong,
  Khurelbaatar, Moroz, Mayfield, and Sayle}]{irwin2020zinc20}
John~J Irwin, Khanh~G Tang, Jennifer Young, Chinzorig Dandarchuluun, Benjamin~R
  Wong, Munkhzul Khurelbaatar, Yurii~S Moroz, John Mayfield, and Roger~A Sayle.
  2020.
\newblock Zinc20—a free ultralarge-scale chemical database for ligand
  discovery.
\newblock \emph{Journal of chemical information and modeling},
  60(12):6065--6073.

\bibitem[{Khurana et~al.(2018)Khurana, Rawi, Kunji, Chuang, Bensmail, and
  Mall}]{DBLP:journals/bioinformatics/KhuranaRKCBM18}
Sameer Khurana, Reda Rawi, Khalid Kunji, Gwo{-}Yu Chuang, Halima Bensmail, and
  Raghvendra Mall. 2018.
\newblock \href {https://doi.org/10.1093/bioinformatics/bty166} {Deepsol: a
  deep learning framework for sequence-based protein solubility prediction}.
\newblock \emph{Bioinform.}, 34(15):2605--2613.

\bibitem[{Kim et~al.(2023)Kim, Chen, Cheng, Gindulyte, He, He, Li, Shoemaker,
  Thiessen, Yu et~al.}]{kim2023pubchem}
Sunghwan Kim, Jie Chen, Tiejun Cheng, Asta Gindulyte, Jia He, Siqian He,
  Qingliang Li, Benjamin~A Shoemaker, Paul~A Thiessen, Bo~Yu, et~al. 2023.
\newblock Pubchem 2023 update.
\newblock \emph{Nucleic Acids Research}, 51(D1):D1373--D1380.

\bibitem[{Kim et~al.(2019)Kim, Thiessen, Cheng, Zhang, Gindulyte, and
  Bolton}]{kim2019pug}
Sunghwan Kim, Paul~A Thiessen, Tiejun Cheng, Jian Zhang, Asta Gindulyte, and
  Evan~E Bolton. 2019.
\newblock Pug-view: programmatic access to chemical annotations integrated in
  pubchem.
\newblock \emph{Journal of cheminformatics}, 11(1):1--11.

\bibitem[{Kipf and Welling(2017)}]{DBLP:conf/iclr/KipfW17}
Thomas~N. Kipf and Max Welling. 2017.
\newblock \href {https://openreview.net/forum?id=SJU4ayYgl} {Semi-supervised
  classification with graph convolutional networks}.
\newblock In \emph{5th International Conference on Learning Representations,
  {ICLR} 2017, Toulon, France, April 24-26, 2017, Conference Track
  Proceedings}. OpenReview.net.

\bibitem[{Krenn et~al.(2022)Krenn, Ai, Barthel, Carson, Frei, Frey, Friederich,
  Gaudin, Gayle, Jablonka et~al.}]{krenn2022selfies}
Mario Krenn, Qianxiang Ai, Senja Barthel, Nessa Carson, Angelo Frei, Nathan~C
  Frey, Pascal Friederich, Th{\'e}ophile Gaudin, Alberto~Alexander Gayle,
  Kevin~Maik Jablonka, et~al. 2022.
\newblock Selfies and the future of molecular string representations.
\newblock \emph{Patterns}, 3(10):100588.

\bibitem[{Krenn et~al.(2020)Krenn, H{\"a}se, Nigam, Friederich, and
  Aspuru-Guzik}]{krenn2020self}
Mario Krenn, Florian H{\"a}se, AkshatKumar Nigam, Pascal Friederich, and Alan
  Aspuru-Guzik. 2020.
\newblock Self-referencing embedded strings (selfies): A 100\% robust molecular
  string representation.
\newblock \emph{Machine Learning: Science and Technology}, 1(4):045024.

\bibitem[{Kudo and Richardson(2018)}]{DBLP:conf/emnlp/KudoR18}
Taku Kudo and John Richardson. 2018.
\newblock \href {https://doi.org/10.18653/v1/d18-2012} {Sentencepiece: {A}
  simple and language independent subword tokenizer and detokenizer for neural
  text processing}.
\newblock In \emph{Proceedings of the 2018 Conference on Empirical Methods in
  Natural Language Processing, {EMNLP} 2018: System Demonstrations, Brussels,
  Belgium, October 31 - November 4, 2018}, pages 66--71. Association for
  Computational Linguistics.

\bibitem[{Landrum(2021)}]{Landrum2021RDKit2021_03_2}
Greg Landrum. 2021.
\newblock \href {https://github.com/rdkit/rdkit/releases/tag/Release_2021_03_2}
  {Rdkit: Open-source cheminformatics software}.
\newblock GitHub release.

\bibitem[{Lee et~al.(2019)Lee, Keum, and Nam}]{Lee2019DeepConvDTIPO}
Ingoo Lee, Jongsoo Keum, and Hojung Nam. 2019.
\newblock {DeepConv-DTI}: Prediction of drug-target interactions via deep
  learning with convolution on protein sequences.
\newblock \emph{PLoS Computational Biology}, 15.

\bibitem[{Lee et~al.(2020)Lee, Yoon, Kim, Kim, Kim, So, and
  Kang}]{lee2020biobert}
Jinhyuk Lee, Wonjin Yoon, Sungdong Kim, Donghyeon Kim, Sunkyu Kim, Chan~Ho So,
  and Jaewoo Kang. 2020.
\newblock Biobert: a pre-trained biomedical language representation model for
  biomedical text mining.
\newblock \emph{Bioinformatics}, 36(4):1234--1240.

\bibitem[{Li et~al.(2023)Li, Liu, Fan, Wei, Liu, Tang, and
  Li}]{li2023empowering}
Jiatong Li, Yunqing Liu, Wenqi Fan, Xiao-Yong Wei, Hui Liu, Jiliang Tang, and
  Qing Li. 2023.
\newblock Empowering molecule discovery for molecule-caption translation with
  large language models: A chatgpt perspective.
\newblock \emph{arXiv preprint arXiv:2306.06615}.

\bibitem[{Lin(2004)}]{lin2004rouge}
Chin-Yew Lin. 2004.
\newblock Rouge: A package for automatic evaluation of summaries.
\newblock In \emph{Text summarization branches out}, pages 74--81.

\bibitem[{Lin et~al.(2022)Lin, Akin, Rao, Hie, Zhu, Lu, dos Santos~Costa,
  Fazel-Zarandi, Sercu, Candido et~al.}]{lin2022language}
Zeming Lin, Halil Akin, Roshan Rao, Brian Hie, Zhongkai Zhu, Wenting Lu, Allan
  dos Santos~Costa, Maryam Fazel-Zarandi, Tom Sercu, Sal Candido, et~al. 2022.
\newblock Language models of protein sequences at the scale of evolution enable
  accurate structure prediction.
\newblock \emph{BioRxiv}.

\bibitem[{Lipman and Pearson(1985)}]{lipman1985rapid}
David~J Lipman and William~R Pearson. 1985.
\newblock Rapid and sensitive protein similarity searches.
\newblock \emph{Science}, 227(4693):1435--1441.

\bibitem[{Lipscomb(2000)}]{lipscomb2000medical}
Carolyn~E Lipscomb. 2000.
\newblock Medical subject headings (mesh).
\newblock \emph{Bulletin of the Medical Library Association}, 88(3):265.

\bibitem[{Liu et~al.(2015)Liu, Sun, Guan, Zheng, and Zhou}]{liu2015improving}
Hui Liu, Jianjiang Sun, Jihong Guan, Jie Zheng, and Shuigeng Zhou. 2015.
\newblock Improving compound--protein interaction prediction by building up
  highly credible negative samples.
\newblock \emph{Bioinformatics}, 31(12):i221--i229.

\bibitem[{Liu et~al.(2022)Liu, Wang, Liu, Lasenby, Guo, and
  Tang}]{DBLP:conf/iclr/LiuWLLGT22}
Shengchao Liu, Hanchen Wang, Weiyang Liu, Joan Lasenby, Hongyu Guo, and Jian
  Tang. 2022.
\newblock \href {https://openreview.net/forum?id=xQUe1pOKPam} {Pre-training
  molecular graph representation with 3d geometry}.
\newblock In \emph{The Tenth International Conference on Learning
  Representations, {ICLR} 2022, Virtual Event, April 25-29, 2022}.
  OpenReview.net.

\bibitem[{Liu et~al.(2023{\natexlab{a}})Liu, Zhu, Lu, Xu, Nie, Gitter, Xiao,
  Tang, Guo, and Anandkumar}]{liu2023text}
Shengchao Liu, Yutao Zhu, Jiarui Lu, Zhao Xu, Weili Nie, Anthony Gitter,
  Chaowei Xiao, Jian Tang, Hongyu Guo, and Anima Anandkumar.
  2023{\natexlab{a}}.
\newblock A text-guided protein design framework.
\newblock \emph{arXiv preprint arXiv:2302.04611}.

\bibitem[{Liu et~al.(2007)Liu, Lin, Wen, Jorissen, and
  Gilson}]{liu2007bindingdb}
Tiqing Liu, Yuhmei Lin, Xin Wen, Robert~N Jorissen, and Michael~K Gilson. 2007.
\newblock Bindingdb: a web-accessible database of experimentally determined
  protein--ligand binding affinities.
\newblock \emph{Nucleic acids research}, 35(suppl\_1):D198--D201.

\bibitem[{Liu et~al.(2019)Liu, Ott, Goyal, Du, Joshi, Chen, Levy, Lewis,
  Zettlemoyer, and Stoyanov}]{liu2019roberta}
Yinhan Liu, Myle Ott, Naman Goyal, Jingfei Du, Mandar Joshi, Danqi Chen, Omer
  Levy, Mike Lewis, Luke Zettlemoyer, and Veselin Stoyanov. 2019.
\newblock Roberta: A robustly optimized bert pretraining approach.
\newblock \emph{arXiv preprint arXiv:1907.11692}.

\bibitem[{Liu et~al.(2023{\natexlab{b}})Liu, Zhang, Xia, Wu, Xie, Qin, Zhang,
  and Liu}]{liu2023molxpt}
Zequn Liu, Wei Zhang, Yingce Xia, Lijun Wu, Shufang Xie, Tao Qin, Ming Zhang,
  and Tie{-}Yan Liu. 2023{\natexlab{b}}.
\newblock \href {https://doi.org/10.18653/v1/2023.acl-short.138} {Molxpt:
  Wrapping molecules with text for generative pre-training}.
\newblock In \emph{Proceedings of the 61st Annual Meeting of the Association
  for Computational Linguistics (Volume 2: Short Papers), {ACL} 2023, Toronto,
  Canada, July 9-14, 2023}, pages 1606--1616. Association for Computational
  Linguistics.

\bibitem[{Loshchilov and Hutter(2019)}]{DBLP:conf/iclr/LoshchilovH19}
Ilya Loshchilov and Frank Hutter. 2019.
\newblock \href {https://openreview.net/forum?id=Bkg6RiCqY7} {Decoupled weight
  decay regularization}.
\newblock In \emph{7th International Conference on Learning Representations,
  {ICLR} 2019, New Orleans, LA, USA, May 6-9, 2019}. OpenReview.net.

\bibitem[{Luo et~al.(2022)Luo, Sun, Xia, Qin, Zhang, Poon, and
  Liu}]{luo2022biogpt}
Renqian Luo, Liai Sun, Yingce Xia, Tao Qin, Sheng Zhang, Hoifung Poon, and
  Tie-Yan Liu. 2022.
\newblock Biogpt: generative pre-trained transformer for biomedical text
  generation and mining.
\newblock \emph{Briefings in Bioinformatics}, 23(6).

\bibitem[{Luo et~al.(2023)Luo, Huang, Hong, Yang, Zhang, Wu, and
  Nie}]{luo2023empowering}
Yizhen Luo, Kui Huang, Massimo Hong, Kai Yang, Jiahuan Zhang, Yushuai Wu, and
  Zaiqin Nie. 2023.
\newblock Empowering ai drug discovery with explicit and implicit knowledge.
\newblock \emph{arXiv preprint arXiv:2305.01523}.

\bibitem[{Medsker and Jain(2001)}]{medsker2001recurrent}
Larry~R Medsker and LC~Jain. 2001.
\newblock Recurrent neural networks.
\newblock \emph{Design and Applications}, 5:64--67.

\bibitem[{Miller et~al.(2009)Miller, Vandome, and
  McBrewster}]{miller2009levenshtein}
Frederic~P Miller, Agnes~F Vandome, and John McBrewster. 2009.
\newblock Levenshtein distance: Information theory, computer science, string
  (computer science), string metric, damerau? levenshtein distance, spell
  checker, hamming distance.

\bibitem[{Nawrot(2023)}]{Nawrot_nanoT5_2023}
Piotr Nawrot. 2023.
\newblock \href {https://doi.org/10.5281/zenodo.7757548} {{nanoT5}}.

\bibitem[{Nguyen et~al.(2021)Nguyen, Le, Quinn, Nguyen, Le, and
  Venkatesh}]{Nguyen2020GraphDTAPD}
Thin Nguyen, Hang Le, T.~Quinn, Tri~Minh Nguyen, Thuc~Duy Le, and Svetha
  Venkatesh. 2021.
\newblock Graph{DTA}: Predicting drug-target binding affinity with graph neural
  networks.
\newblock \emph{Bioinformatics}, 37(8):1140--1147.

\bibitem[{O'Boyle and Dalke(2018)}]{o2018deepsmiles}
Noel O'Boyle and Andrew Dalke. 2018.
\newblock Deepsmiles: an adaptation of smiles for use in machine-learning of
  chemical structures.

\bibitem[{O'Shea and Nash(2015)}]{o2015introduction}
Keiron O'Shea and Ryan Nash. 2015.
\newblock An introduction to convolutional neural networks.
\newblock \emph{arXiv preprint arXiv:1511.08458}.

\bibitem[{Pan et~al.(2010)Pan, Zhang, and Shen}]{pan2010large}
Xiao-Yong Pan, Ya-Nan Zhang, and Hong-Bin Shen. 2010.
\newblock Large-scale prediction of human protein- protein interactions from
  amino acid sequence based on latent topic features.
\newblock \emph{Journal of proteome research}, 9(10):4992--5001.

\bibitem[{Papineni et~al.(2002)Papineni, Roukos, Ward, and
  Zhu}]{DBLP:conf/acl/PapineniRWZ02}
Kishore Papineni, Salim Roukos, Todd Ward, and Wei{-}Jing Zhu. 2002.
\newblock \href {https://doi.org/10.3115/1073083.1073135} {Bleu: a method for
  automatic evaluation of machine translation}.
\newblock In \emph{Proceedings of the 40th Annual Meeting of the Association
  for Computational Linguistics, July 6-12, 2002, Philadelphia, PA, {USA}},
  pages 311--318. {ACL}.

\bibitem[{Pearson and Lipman(1988)}]{pearson1988improved}
William~R Pearson and David~J Lipman. 1988.
\newblock Improved tools for biological sequence comparison.
\newblock \emph{Proceedings of the National Academy of Sciences},
  85(8):2444--2448.

\bibitem[{Peri et~al.(2003)Peri, Navarro, Amanchy, Kristiansen, Jonnalagadda,
  Surendranath, Niranjan, Muthusamy, Gandhi, Gronborg
  et~al.}]{peri2003development}
Suraj Peri, J~Daniel Navarro, Ramars Amanchy, Troels~Z Kristiansen,
  Chandra~Kiran Jonnalagadda, Vineeth Surendranath, Vidya Niranjan, Babylakshmi
  Muthusamy, TKB Gandhi, Mads Gronborg, et~al. 2003.
\newblock Development of human protein reference database as an initial
  platform for approaching systems biology in humans.
\newblock \emph{Genome research}, 13(10):2363--2371.

\bibitem[{Preuer et~al.(2018)Preuer, Renz, Unterthiner, Hochreiter, and
  Klambauer}]{DBLP:journals/jcisd/PreuerRUHK18}
Kristina Preuer, Philipp Renz, Thomas Unterthiner, Sepp Hochreiter, and
  G{\"{u}}nter Klambauer. 2018.
\newblock \href {https://doi.org/10.1021/acs.jcim.8b00234} {Fr{\'{e}}chet
  chemnet distance: {A} metric for generative models for molecules in drug
  discovery}.
\newblock \emph{J. Chem. Inf. Model.}, 58(9):1736--1741.

\bibitem[{Radford et~al.(2018)Radford, Narasimhan, Salimans, Sutskever
  et~al.}]{radford2018improving}
Alec Radford, Karthik Narasimhan, Tim Salimans, Ilya Sutskever, et~al. 2018.
\newblock Improving language understanding by generative pre-training.

\bibitem[{Radford et~al.(2019)Radford, Wu, Child, Luan, Amodei, Sutskever
  et~al.}]{radford2019language}
Alec Radford, Jeffrey Wu, Rewon Child, David Luan, Dario Amodei, Ilya
  Sutskever, et~al. 2019.
\newblock Language models are unsupervised multitask learners.
\newblock \emph{OpenAI blog}, 1(8):9.

\bibitem[{Raffel et~al.(2020)Raffel, Shazeer, Roberts, Lee, Narang, Matena,
  Zhou, Li, and Liu}]{raffel2020exploring}
Colin Raffel, Noam Shazeer, Adam Roberts, Katherine Lee, Sharan Narang, Michael
  Matena, Yanqi Zhou, Wei Li, and Peter~J Liu. 2020.
\newblock Exploring the limits of transfer learning with a unified text-to-text
  transformer.
\newblock \emph{The Journal of Machine Learning Research}, 21(1):5485--5551.

\bibitem[{Rives et~al.(2021)Rives, Meier, Sercu, Goyal, Lin, Liu, Guo, Ott,
  Zitnick, Ma et~al.}]{rives2021biological}
Alexander Rives, Joshua Meier, Tom Sercu, Siddharth Goyal, Zeming Lin, Jason
  Liu, Demi Guo, Myle Ott, C~Lawrence Zitnick, Jerry Ma, et~al. 2021.
\newblock Biological structure and function emerge from scaling unsupervised
  learning to 250 million protein sequences.
\newblock \emph{Proceedings of the National Academy of Sciences},
  118(15):e2016239118.

\bibitem[{Robertson and Zaragoza(2009)}]{DBLP:journals/ftir/RobertsonZ09}
Stephen~E. Robertson and Hugo Zaragoza. 2009.
\newblock \href {https://doi.org/10.1561/1500000019} {The probabilistic
  relevance framework: {BM25} and beyond}.
\newblock \emph{Found. Trends Inf. Retr.}, 3(4):333--389.

\bibitem[{Rogers and Hahn(2010{\natexlab{a}})}]{rogers2010extended}
David Rogers and Mathew Hahn. 2010{\natexlab{a}}.
\newblock Extended-connectivity fingerprints.
\newblock \emph{Journal of chemical information and modeling}, 50(5):742--754.

\bibitem[{Rogers and Hahn(2010{\natexlab{b}})}]{DBLP:journals/jcisd/RogersH10}
David Rogers and Mathew Hahn. 2010{\natexlab{b}}.
\newblock \href {https://doi.org/10.1021/ci100050t} {Extended-connectivity
  fingerprints}.
\newblock \emph{J. Chem. Inf. Model.}, 50(5):742--754.

\bibitem[{Rong et~al.(2020)Rong, Bian, Xu, Xie, Wei, Huang, and
  Huang}]{rong2020self}
Yu~Rong, Yatao Bian, Tingyang Xu, Weiyang Xie, Ying Wei, Wenbing Huang, and
  Junzhou Huang. 2020.
\newblock Self-supervised graph transformer on large-scale molecular data.
\newblock \emph{Advances in Neural Information Processing Systems},
  33:12559--12571.

\bibitem[{Saravanan and Gautham(2015)}]{saravanan2015harnessing}
Vijayakumar Saravanan and Namasivayam Gautham. 2015.
\newblock Harnessing computational biology for exact linear b-cell epitope
  prediction: a novel amino acid composition-based feature descriptor.
\newblock \emph{Omics: a journal of integrative biology}, 19(10):648--658.

\bibitem[{Schneider et~al.(2015)Schneider, Sayle, and
  Landrum}]{DBLP:journals/jcisd/SchneiderSL15}
Nadine Schneider, Roger~A. Sayle, and Gregory~A. Landrum. 2015.
\newblock \href {https://doi.org/10.1021/acs.jcim.5b00543} {Get your atoms in
  order - an open-source implementation of a novel and robust molecular
  canonicalization algorithm}.
\newblock \emph{J. Chem. Inf. Model.}, 55(10):2111--2120.

\bibitem[{Steinegger and S{\"o}ding(2018)}]{steinegger2018clustering}
Martin Steinegger and Johannes S{\"o}ding. 2018.
\newblock Clustering huge protein sequence sets in linear time.
\newblock \emph{Nature communications}, 9(1):2542.

\bibitem[{Sterling and Irwin(2015)}]{DBLP:journals/jcisd/SterlingI15}
Teague Sterling and John~J. Irwin. 2015.
\newblock \href {https://doi.org/10.1021/acs.jcim.5b00559} {{ZINC} 15 - ligand
  discovery for everyone}.
\newblock \emph{J. Chem. Inf. Model.}, 55(11):2324--2337.

\bibitem[{Su et~al.(2022)Su, Du, Yang, Zhou, Li, Rao, Sun, Lu, and
  Wen}]{su2022molecular}
Bing Su, Dazhao Du, Zhao Yang, Yujie Zhou, Jiangmeng Li, Anyi Rao, Hao Sun,
  Zhiwu Lu, and Ji-Rong Wen. 2022.
\newblock A molecular multimodal foundation model associating molecule graphs
  with natural language.
\newblock \emph{arXiv preprint arXiv:2209.05481}.

\bibitem[{Sung et~al.(2022)Sung, Jeong, Choi, Kim, Lee, and
  Kang}]{sung2022bern2}
Mujeen Sung, Minbyul Jeong, Yonghwa Choi, Donghyeon Kim, Jinhyuk Lee, and
  Jaewoo Kang. 2022.
\newblock Bern2: an advanced neural biomedical named entity recognition and
  normalization tool.
\newblock \emph{Bioinformatics}, 38(20):4837--4839.

\bibitem[{Suzek et~al.(2007)Suzek, Huang, McGarvey, Mazumder, and
  Wu}]{suzek2007uniref}
Baris~E Suzek, Hongzhan Huang, Peter McGarvey, Raja Mazumder, and Cathy~H Wu.
  2007.
\newblock Uniref: comprehensive and non-redundant uniprot reference clusters.
\newblock \emph{Bioinformatics}, 23(10):1282--1288.

\bibitem[{Taylor et~al.(2022)Taylor, Kardas, Cucurull, Scialom, Hartshorn,
  Saravia, Poulton, Kerkez, and Stojnic}]{taylor2022galactica}
Ross Taylor, Marcin Kardas, Guillem Cucurull, Thomas Scialom, Anthony
  Hartshorn, Elvis Saravia, Andrew Poulton, Viktor Kerkez, and Robert Stojnic.
  2022.
\newblock Galactica: A large language model for science.
\newblock \emph{arXiv preprint arXiv:2211.09085}.

\bibitem[{Vaswani et~al.(2017)Vaswani, Shazeer, Parmar, Uszkoreit, Jones,
  Gomez, Kaiser, and Polosukhin}]{vaswani2017attention}
Ashish Vaswani, Noam Shazeer, Niki Parmar, Jakob Uszkoreit, Llion Jones,
  Aidan~N Gomez, {\L}ukasz Kaiser, and Illia Polosukhin. 2017.
\newblock Attention is all you need.
\newblock \emph{Advances in neural information processing systems}, 30.

\bibitem[{Wang et~al.(2022)Wang, Wang, Cao, and
  Barati~Farimani}]{wang2022molecular}
Yuyang Wang, Jianren Wang, Zhonglin Cao, and Amir Barati~Farimani. 2022.
\newblock Molecular contrastive learning of representations via graph neural
  networks.
\newblock \emph{Nature Machine Intelligence}, 4(3):279--287.

\bibitem[{Wei et~al.(2022)Wei, Bosma, Zhao, Guu, Yu, Lester, Du, Dai, and
  Le}]{DBLP:conf/iclr/WeiBZGYLDDL22}
Jason Wei, Maarten Bosma, Vincent~Y. Zhao, Kelvin Guu, Adams~Wei Yu, Brian
  Lester, Nan Du, Andrew~M. Dai, and Quoc~V. Le. 2022.
\newblock \href {https://openreview.net/forum?id=gEZrGCozdqR} {Finetuned
  language models are zero-shot learners}.
\newblock In \emph{The Tenth International Conference on Learning
  Representations, {ICLR} 2022, Virtual Event, April 25-29, 2022}.
  OpenReview.net.

\bibitem[{Weininger(1988)}]{weininger1988smiles}
David Weininger. 1988.
\newblock Smiles, a chemical language and information system. 1. introduction
  to methodology and encoding rules.
\newblock \emph{Journal of chemical information and computer sciences},
  28(1):31--36.

\bibitem[{Weininger et~al.(1989)Weininger, Weininger, and
  Weininger}]{weininger1989smiles}
David Weininger, Arthur Weininger, and Joseph~L Weininger. 1989.
\newblock Smiles. 2. algorithm for generation of unique smiles notation.
\newblock \emph{Journal of chemical information and computer sciences},
  29(2):97--101.

\bibitem[{Wishart et~al.(2018)Wishart, Feunang, Guo, Lo, Marcu, Grant, Sajed,
  Johnson, Li, Sayeeda et~al.}]{wishart2018drugbank}
David~S Wishart, Yannick~D Feunang, An~C Guo, Elvis~J Lo, Ana Marcu, Jason~R
  Grant, Tanvir Sajed, Daniel Johnson, Carin Li, Zinat Sayeeda, et~al. 2018.
\newblock Drugbank 5.0: a major update to the drugbank database for 2018.
\newblock \emph{Nucleic acids research}, 46(D1):D1074--D1082.

\bibitem[{Wu et~al.(2018)Wu, Ramsundar, Feinberg, Gomes, Geniesse, Pappu,
  Leswing, and Pande}]{wu2018moleculenet}
Zhenqin Wu, Bharath Ramsundar, Evan~N Feinberg, Joseph Gomes, Caleb Geniesse,
  Aneesh~S Pappu, Karl Leswing, and Vijay Pande. 2018.
\newblock Moleculenet: a benchmark for molecular machine learning.
\newblock \emph{Chemical science}, 9(2):513--530.

\bibitem[{Xu et~al.(2023{\natexlab{a}})Xu, Woicik, Poon, Altman, and
  Wang}]{xu2023multilingual}
Hanwen Xu, Addie Woicik, Hoifung Poon, Russ~B Altman, and Sheng Wang.
  2023{\natexlab{a}}.
\newblock Multilingual translation for zero-shot biomedical classification
  using biotranslator.
\newblock \emph{Nature Communications}, 14(1):738.

\bibitem[{Xu et~al.(2023{\natexlab{b}})Xu, Yuan, Miret, and
  Tang}]{xu2023protst}
Minghao Xu, Xinyu Yuan, Santiago Miret, and Jian Tang. 2023{\natexlab{b}}.
\newblock Protst: Multi-modality learning of protein sequences and biomedical
  texts.
\newblock \emph{arXiv preprint arXiv:2301.12040}.

\bibitem[{Xu et~al.(2022)Xu, Zhang, Lu, Zhu, Zhang, Chang, Liu, and
  Tang}]{xu2022peer}
Minghao Xu, Zuobai Zhang, Jiarui Lu, Zhaocheng Zhu, Yangtian Zhang, Ma~Chang,
  Runcheng Liu, and Jian Tang. 2022.
\newblock Peer: a comprehensive and multi-task benchmark for protein sequence
  understanding.
\newblock \emph{Advances in Neural Information Processing Systems},
  35:35156--35173.

\bibitem[{Zeng et~al.(2022)Zeng, Yao, Liu, and Sun}]{zeng2022deep}
Zheni Zeng, Yuan Yao, Zhiyuan Liu, and Maosong Sun. 2022.
\newblock A deep-learning system bridging molecule structure and biomedical
  text with comprehension comparable to human professionals.
\newblock \emph{Nature communications}, 13(1):862.

\bibitem[{Zhang et~al.(2021)Zhang, Liu, Wang, Lu, and Lee}]{zhang2021motif}
Zaixi Zhang, Qi~Liu, Hao Wang, Chengqiang Lu, and Chee-Kong Lee. 2021.
\newblock Motif-based graph self-supervised learning for molecular property
  prediction.
\newblock \emph{Advances in Neural Information Processing Systems},
  34:15870--15882.

\bibitem[{Zitnik et~al.(2018)Zitnik, Sosic, and Leskovec}]{zitnik2018biosnap}
Marinka Zitnik, Rok Sosic, and Jure Leskovec. 2018.
\newblock Biosnap datasets: Stanford biomedical network dataset collection.
\newblock \emph{Note: http://snap. stanford. edu/biodata Cited by}, 5(1).

\end{thebibliography}
\bibliographystyle{acl_natbib}

\clearpage
\appendix

\section{Reproducibility}
The codes for our \method{} are available at \url{https://github.com/QizhiPei/BioT5}. 

\section{NER and Entity Linking Process}
\label{sec:ner_el}
We follow KV-PLM~\citep{zeng2022deep} and MolXPT~\citep{liu2023molxpt} to conduct Named Entity Recognition (NER) and Entity Linking for the bio-entity names appearing in the scientific text.
More specifically, we firstly utilize BERN2~\citep{sung2022bern2}, an advanced neural Named Entity Recognition (NER) tool in biomedical fields, to identify all instances of molecule or protein mentions. 
Subsequently, we map them to corresponding entities within publicly accessible knowledge bases.
For molecule, we use ChEBI~\citep{hastings2016chebi} and MeSH~\citep{lipscomb2000medical} database, and for protein we use NCBI Gene~\citep{brister2015ncbi} database.
Then we can get the corresponding molecule SELFIES and protein FASTA for the matched entities.
\begin{figure}[h]
    \centering
    % \vspace{-0.7cm}
    \includegraphics[width=\linewidth]{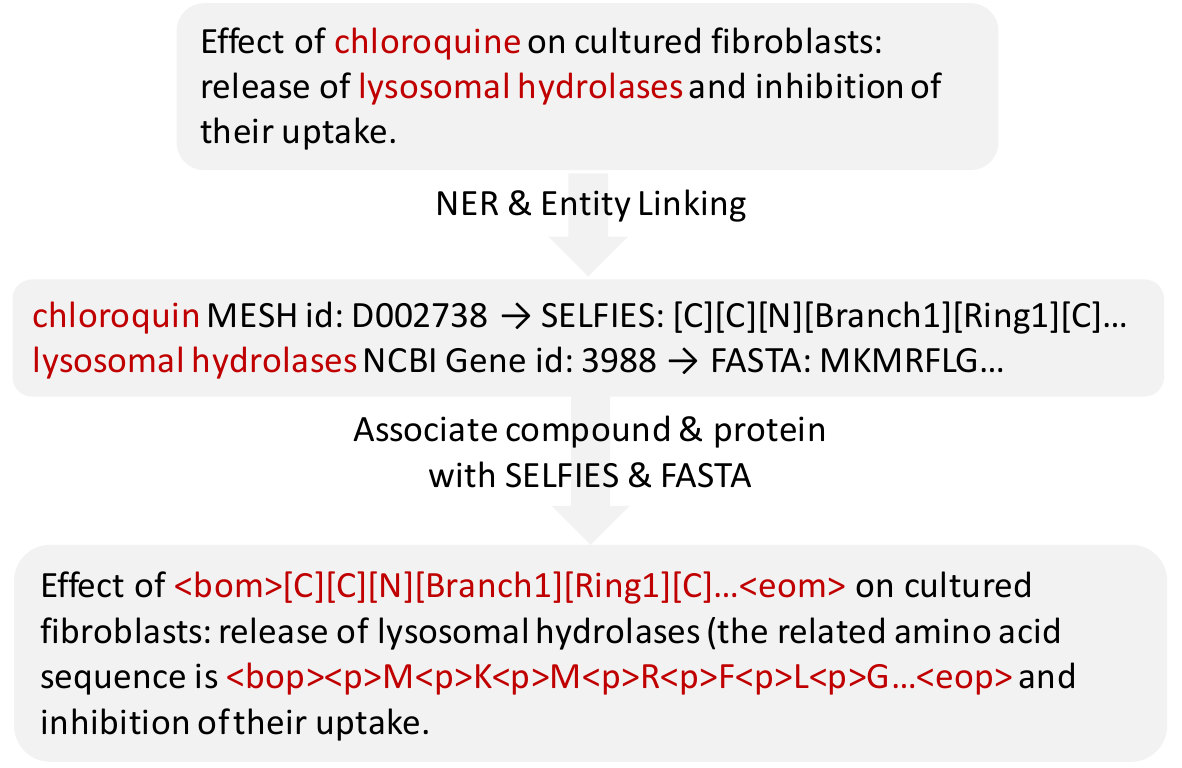}
    \caption{Wrapped text matching and mapping process.}
    \label{fig:ner}
    % \vspace{-0.52cm}
\end{figure}
As shown in Figure~\ref{fig:ner}, for molecule, we directly replace all the detected names with its SELFIES string;
for protein, due to the length limitation, if a sentence consists of more than one protein entity, we only randomly choose one to append the protein FASTA to the name.
The motivation for appending protein FASTA instead of replacing is that the genes are transcribed and translated to generate proteins. 
Therefore, unlike the molecule names directly representing the molecule, the relation between gene names and protein FASTA is indirect.
Note that the replacement or appendage will not happen in every sentence.
Only those with detected bio-entities will be done the above process.

\section{Dictionary and SELFIES Conversion}
\label{sec:dict_and_conversion}
For molecule-related datasets, when only SMILES is provided, we utilize \textit{selfies\footnote{\url{https://github.com/aspuru-guzik-group/selfies}}} package to convert SMILES into SELFIES.

\section{Molecule-Text Generation Metrics}
\label{sec:mol_text_metric}
We follow~\citet{DBLP:conf/emnlp/EdwardsLRHCJ22} to use the same evaluation metrics for molecule captioning and text-based molecule generation tasks.
To ensure a fair comparison, we convert the molecule SEIFLES to SMILES before calculating these metrics.
\subsection{Molecule Captioning Metrics}
In the molecule caption task, NLP metrics like BLEU~\citep{DBLP:conf/acl/PapineniRWZ02}, ROUGE~\citep{lin2004rouge}, and METEOR~\citep{DBLP:conf/acl/BanerjeeL05} are utilized to evaluate the closeness of the generated description to the ground truth description.
We also adopt \textit{Text2Mol} metric, which is proposed by~\citet{DBLP:conf/emnlp/EdwardsZJ21} and employ pre-trained models to measure the similarity between the description and ground truth molecule.
Higher similarity means that the given text description is more relevant to the molecule, and the Text2Mol score between the ground truth description and molecule is also computed for comparison.

\subsection{Text-based Molecule Generation Metrics}
Since molecules can be represented in bio-sequence structure, NLP metrics like BLEU~\citep{DBLP:conf/acl/PapineniRWZ02} and Exact Match scores between generated and ground truth SMILES are directly applied for evaluation.
Additionally, we also report performance on molecule-specific metrics: three molecule fingerprints (FTS) similarity scores-MACCS~\citep{durant2002reoptimization}, RDK~\citep{DBLP:journals/jcisd/SchneiderSL15}, and Morgan~\citep{rogers2010extended};
Levenshtein distance~\citep{miller2009levenshtein}; 
FCD score~\citep{DBLP:journals/jcisd/PreuerRUHK18}, which measures molecule similarities according to biological information based on pre-trained ``ChemNet'';
validity, which is the percentage of the valid SMILES that can be processed by RDKit~\citep{Landrum2021RDKit2021_03_2}.
The \textit{Text2Mol} metric is also used to measure the similarity between the molecule SMILES and ground truth description.

\section{Pre-training Details}
\label{sec:pretrain_dataset}
\subsection{Special Tokens}
\label{sec:special_tokens}
In the pre-training of \method{}, we conduct translation tasks on molecule-text pairs and protein-text pairs extracted from PubChem~\citep{kim2023pubchem} and Swiss-Prot~\citep{boutet2007uniprotkb} separately.
We format the text description from these database entries using special tokens, which serve as anchors for embedding scientific context and structure.
For molecule, we use \textit{MOLECULE NAME} and \textit{DESCRIPTION} to represent its name and description including properties, functions, etc.
For protein, similar to~\citet{xu2023protst}, we use \textit{PROTEIN NAME}, \textit{FUNCTION}, \textit{SUBCELLULAR LOCATION}, and \textit{PROTEIN FAMILIES} to represent its name, functions, location and topology in the cell, and families it belongs to.
A complete text description is created by concatenating these fields sequentially, omitting any missing fields.
Through special tokens, we can effectively encode the intricate information associated with each bio-entity.
\subsection{Hyper-parameters}
We use the codebase {\em nanoT5}~\citep{Nawrot_nanoT5_2023} for \method{} pre-training.
We pre-train \method{} for $350$K steps on eight NVIDIA 80GB A100 GPUs.
The batch size is $96$ per GPU, in which a batch includes six types of data.
The ``translation'' directions for molecule-text and protein-text pair are randomly selected for each sample with a probability of $0.5$.
We use AdamW~\citep{DBLP:conf/iclr/LoshchilovH19} with Root Mean Square (RMS) scaling Optimizer for optimization.
The learning rate scheduler is cosine annealing with the base learning rate set to $1e-2$ and the minimum learning rate set to $1e-5$. 
The number of warm-up steps is 10,000 and the dropout rate is $0.0$.
The maximum input length for pre-training is $512$. 
Unlike absolute position encodings, T5~\citep{raffel2020exploring} use relative position encodings.
This makes the model flexible to inputs of different lengths, which is helpful for downstream fine-tuning.

\section{Fine-tuning Details}
\label{sec:finetune_detail}
\begin{table*}[t]
    \centering
    \resizebox{0.9\textwidth}{!}{
    \small
    \begin{tabular}{ccccc}
        \toprule
        \bf{Task/Dataset} & \bf{Task Type} & \bf{\#Train} & \bf{\#Validation} & \bf{\#Test} \\
        \midrule
        \multicolumn{5}{c}{\bf{Molecule Property Prediction}} \\
        \midrule
        \bf{BBBP} & Molecule-wise Classification & 1,631 & 204 & 204 \\
        \bf{Tox21} & Molecule-wise Classification & 6,264 & 783 & 784 \\
        \bf{ClinTox} & Molecule-wise Classification & 1,181 & 148 & 148 \\
        \bf{HIV} & Molecule-wise Classification & 32,901 & 4,113 & 4,113 \\
        \bf{BACE} & Molecule-wise Classification & 1,210 & 151 & 152 \\
        \bf{SIDER} & Molecule-wise Classification & 1,141 & 143 & 143 \\
        \midrule
        \multicolumn{5}{c}{\bf{Protein Property Prediction}} \\
        \midrule
        \bf{Solubility prediction} & Protein-wise Classification & 62,478 & 1,999 & 1,999\\
        \bf{Localization prediction} & Protein-wise Classification & 5,184 & 1,749 & 1,749 \\
        \midrule
        \multicolumn{5}{c}{\bf{Drug-target Interaction Prediction}} \\
        \midrule
        \bf{BioSNAP} & Molecule-protein Classification & 19,224 & 2,747 & 5,493 \\
        \bf{Human} & Molecule-protein Classification & 4,197 & 600 & 1,200 \\
        % \bf{BindingDB} & Molecule-protein Classification & 50,149 & 5,604 & 5,505 \\
        \bf{BindingDB} & Molecule-protein Classification & 34,439 & 4,920 & 9,840 \\
        \midrule
        \multicolumn{5}{c}{\bf{Protein-protein Interaction Prediction}} \\
        \midrule
        \bf{Yeast} & Protein-pair Classification & 4,945 & 394 & 394 \\
        \bf{Human} & Protein-pair Classification & 35,669 & 237 & 237 \\
        \midrule
        \multicolumn{5}{c}{\bf{Molecule Captioning and Text-based Molecule Generation}} \\
        \midrule
        \bf{ChEBI-20} & Molecule-text Translation & 26,407 & 3,301 & 3,300 \\
        \bottomrule
    \end{tabular}
    }
    \caption{Downstream task descriptions, including task or dataset name, type, and the size of each split.} 
    \label{tab:downstream_task}
\end{table*}
In this section, we provide details about downstream tasks, including datasets, compared baselines, and prompts.
Some statistics about downstream tasks are shown in Table~\ref{tab:downstream_task}
When displaying prompts, \selfies{} refers to the molecule SELFIES, and \fasta{} refers to the protein FASTA. 

\subsection{Single-instance Prediction}
\subsubsection{Molecule Property Prediction}
All the datasets are split using an $8:1:1$ ratio for train, validation, and test, respectively.
We use the scaffold splitting method, in which molecules are categorized according to the Bemis-Murcko scaffold representation.

\noindent{\textbf{Datasets}}

\noindent(1) The BBBP (Blood-Brain Barrier Penetration) is curated with the intention of aiding the modeling and forecasting of barrier permeability. It comprises compounds that are categorized using binary labels, indicating whether they can penetrate the blood-brain barrier.

\noindent(2) The Tox21 ("Toxicology in the 21st Century") initiative established a publicly accessible database that quantifies the toxicity levels of various compounds. The dataset encompasses qualitative toxicity assessments (binary labels) for approximately 8,000 compounds, targeting 12 distinct biological pathways such as nuclear receptors and stress response mechanisms.

\noindent(3) The ClinTox dataset contrasts FDA-approved drugs with those that have been unsuccessful in clinical trials owing to toxicity issues. This dataset incorporates two classification objectives for 1,491 drug compounds with established chemical structures:
(i) Presence or absence of toxicity in clinical trials; (ii) approved or unapproved by FDA.

\noindent(4) The HIV dataset assesses the inhibitory potential of over 40,000 compounds on HIV replication. The screening outcomes were classified into three categories: Confirmed Inactive (CI), Confirmed Active (CA), and Confirmed Moderately Active (CM). Subsequently, the latter two labels were combined, transforming the task into a binary classification between inactive (CI) and active (CA and CM) categories.

\noindent(5) The BACE dataset presents quantitative IC50 values and qualitative binary labels for a collection of inhibitors targeting human beta-secretase 1 (BACE-1).

\noindent(6) The SIDER (Side Effect Resource) is a comprehensive database that consists of marketed drugs and their corresponding adverse drug reactions (ADR). The drug side effects in SIDER are organized into 27 system organ classes, adhering to the MedDRA classifications. This dataset encompasses data for 1,427 approved drugs.

\noindent{\textbf{Baselines}}

\noindent(1) GROVER~\citep{rong2020self} incorporates Message Passing Networks within a Transformer-style architecture and is pre-trained on large-scale molecular dataset without any supervision. G-Contextual and G-Motif are two variants of GROVER, which are pre-trained on contextual property prediction task and motif prediction task, respectively.

\noindent(2) GraphMVP~\citep{DBLP:conf/iclr/LiuWLLGT22} employs self-supervised learning by capitalizing on the correspondence and consistency between molecule 2D topological structures and 3D geometric views. 

\noindent(3) MGSSL~\citep{zhang2021motif} incorporates a novel self-supervised motif generation framework for Graph Neural Networks.

\noindent(4) MolCLR~\citep{wang2022molecular} is a self-supervised learning framework that capitalizes on substantial unlabelled unique molecules (approximately 10 million)

\noindent(5) GEM~\citep{fang2022geometry} features a specially designed geometry-based graph neural network architecture and several dedicated geometry-level self-supervised learning strategies to capture molecular geometry knowledge effectively.

\noindent(6) KV-PLM~\citep{zeng2022deep} is a BERT-based model designed for molecular representation learning, in which molecule SMILES are appended after its name during the pre-training process. This combination of molecular names and SMILES sequences allows the model to capture both textual and structural information, thereby enhancing its performance in various downstream tasks.

\noindent(7) Galactica~\citep{taylor2022galactica} is a large GPT-based language model which is pre-trained on various corpus like papers, codes, SMILES, protein sequences, etc.

\noindent(8) MoMu~\citep{su2022molecular} is pre-trained using molecular graphs and their semantically related textual data through contrastive learning.

\noindent(9) MolXPT~\cite{liu2023molxpt} is a unified GPT-based language model for text and molecules pre-trained on ``wrapped'' text, where molecule names are replaced with corresponding SMILES.

\noindent{\textbf{Prompts}}

\noindent For the six MoleculeNet datasets mentioned above, the prompts only differ in the \underline{Task Definition}. Therefore, we will only provide the \underline{Instruction} and \underline{Output} for the first dataset, and the remaining datasets will follow the same format.

\noindent(1) BBBP

\noindent{\underline{Task Definition}}:
\textit{Definition: Molecule property prediction task (a binary classification task) for the BBBP dataset. The blood-brain barrier penetration (BBBP) dataset is designed for the modeling and prediction of barrier permeability. If the given molecule can penetrate the blood-brain barrier, indicate via "Yes". Otherwise, response via "No".} 

\noindent{\underline{Instruction}}:
\textit{Now complete the following example - Input: Molecule: \bom{}\selfies{}\eom{} Output:}.

\noindent{\underline{Output}}: \textit{Yes} for inhibitor and \textit{No} instead.

\noindent(2) Tox21

\noindent{\underline{Task Definition}}:
\textit{Definition: Molecule property prediction task (a binary classification task) for the Tox21 dataset. The Tox21 dataset contains qualitative toxicity measurements for 8k compounds on 12 different targets, including nuclear receptors and stress response pathways. If the given molecule can activate/change/affect $\langle$\texttt{target}$\rangle$, indicate via "Yes". Otherwise, response via "No".} where $\langle$\texttt{target}$\rangle$ represents the corresponding receptor, domain, element, gene, potential, or pathway for each subtask.

\noindent(3) ClinTox

\noindent{\underline{Task Definition}}:
\textit{Definition: Molecule property prediction task (a binary classification task) for the ClinTox dataset. The ClinTox dataset compares drugs approved by the FDA and drugs that have failed clinical trials for toxicity reasons. If the given molecule is $\langle$\texttt{Subtask}$\rangle$, indicate via "Yes". Otherwise, response via "No".} where the $\langle$\texttt{Subtask}$\rangle$ is either \textit{toxic} or \textit{FDA approved}.

\noindent(4) HIV

\noindent{\underline{Task Definition}}:
\textit{Definition: Molecule property prediction task (a binary classification task) for the HIV dataset. The HIV dataset was introduced by the Drug Therapeutics Program (DTP) AIDS Antiviral Screen, which tested the ability to inhibit HIV replication for over 40,000 compounds. If the given molecule can inhibit HIV replication, indicate via "Yes". Otherwise, response via "No".}

\noindent(5) BACE

\noindent{\underline{Task Definition}}:
\textit{Definition: Molecule property prediction task (a binary classification task) for the BACE dataset. The BACE dataset provides qualitative (binary label) binding results for a set of inhibitors of human beta-secretase 1 (BACE-1). If the given molecule can inhibit BACE-1, indicate via "Yes". Otherwise, response via "No".}

\noindent(6) SIDER

\noindent{\underline{Task Definition}}:
\textit{Definition: Molecule property prediction task (a binary classification task) for the SIDER dataset. The Side Effect Resource (SIDER) is a dataset of marketed drugs and adverse drug reactions (ADR). If the given molecule can cause the side effect of $\langle$\texttt{side effect}$\rangle$, indicate via "Yes". Otherwise, response via "No".} where $\langle$\texttt{side effect}$\rangle$ refers to the corresponding side effects for each subtask.

\subsubsection{Protein Property Prediction}
\label{sec:protein_property_detail}
\noindent{\textbf{Datasets}}

\noindent(1) Solubility prediction is to predict whether a protein is soluble or not. We follow the same splitting method with DeepSol~\citep{DBLP:journals/bioinformatics/KhuranaRKCBM18}.

\noindent(2) Localization prediction aims predict whether a protein is ``membrane-bound'' or ``soluble'', which is a simple version of subcellular localization prediction task. We follow the same splitting method with DeepLoc~\citep{DBLP:journals/bioinformatics/ArmenterosSSNW17}.

\noindent{\textbf{Baselines}}

\noindent(1) Feature engineers. The DDE (Dipeptide Deviation from Expected Mean)~\citep{saravanan2015harnessing} feature descriptor, consisting of 400 dimensions, is based on the dipeptide frequency within a protein sequence.
The Moran feature descriptor (Moran correlation)~\citep{feng2000prediction}, with 240 dimensions, characterizes the distribution of amino acid properties within a protein sequence. 

\noindent(2) Protein sequence encoders, including LSTM~\citep{hochreiter1997long}, Transformers~\citep{vaswani2017attention}, CNN~\citep{o2015introduction} and ResNet~\citep{he2016deep}.
The amino acid features in the last layer are aggregated for final prediction.

\noindent(3) Pre-trained protein language models.
ProtBert~\citep{elnaggar2021prottrans} and ESM-1b~\citep{rives2021biological} are both pre-trained on a massive dataset of protein sequences using the masked language modeling (MLM) objective. 
Specifically, ProtBert is pre-trained on $2.1$ billion protein sequences obtained from the BFD database~\citep{steinegger2018clustering}, while ESM-1b is pre-trained on a smaller dataset of $24$ million protein sequences sourced from UniRef50~\citep{suzek2007uniref}.

\noindent{\textbf{Prompts}}

\noindent(1) Solubility prediction

\noindent{\underline{Task Definition}}:
\textit{Protein solubility prediction task (a binary classification task) for the solubility dataset. If the given protein is soluble, indicate via "Yes". Otherwise, response via "No".}

\noindent{\underline{Instruction}}
\textit{Now complete the following example - Input: Protein: \bop{}\fasta{}\eop{} Output:}.

\noindent{\underline{Output}}:
\textit{Yes} for soluble protein or \textit{No} instead.

\noindent(2) Localization prediction

\noindent{\underline{Task Definition}}:
\textit{Protein subcellular localization task (a binary classification task). If the given protein is membrane-bound, indicate via "Yes". Otherwise (the protein is soluble), response via "No".}

\noindent{\underline{Instruction}}
\textit{Now complete the following example - Input: Protein: \bop{}\fasta{}\eop{} Output:}.

\noindent{\underline{Output}}:
\textit{Yes} for membrane-bound protein or \textit{No} for soluble protein.

\subsection{Multi-instance Prediction}
\subsubsection{Drug-target Interaction Prediction}
\noindent{\textbf{Datasets}}

\noindent(1) BioSNAP~\citep{zitnik2018biosnap} is derived from the DrugBank database~\citep{wishart2018drugbank} and was created by~\citet{Huang2021MolTransMI} and~\citet{zitnik2018biosnap}. It consists of 4,510 drugs and 2,181 proteins. This dataset is balanced, containing both validated positive interactions and an equal number of randomly selected negative samples from unseen pairs.

\noindent(2) BindingDB~\citep{liu2007bindingdb} is an accessible online database that contains experimentally validated binding affinities. Its main focus is on the interactions between small drug-like molecules and proteins. We follow~\citet{bai2023interpretable} to use a modified version of the BindingDB dataset, which was previously constructed by~\citet{bai2021hierarchical} with reduced bias. 

\noindent(3) Human~\cite{liu2015improving,chen2020transformercpi} is constructed with the inclusion of highly credible negative samples. Following~\citet{bai2023interpretable}, we also use a balanced version of the Human dataset, which contains an equal number of positive and negative samples. 

\noindent{\textbf{Baselines}}

\noindent We compare the performance of \method{} with the following six models on DTI task.

\noindent(1) Support Vector Machine~\citep{cortes1995support} (SVM) on the concatenated fingerprint ECFP4~\citep{DBLP:journals/jcisd/RogersH10} (extended connectivity fingerprint, up to four bonds) and PSC~\citep{DBLP:journals/bioinformatics/CaoXL13} (pseudo-amino acid composition) features.

\noindent(2) Random Forest~\citep{ho1995random} (RF) on the concatenated fingerprint ECFP4 and PSC features.

\noindent(3) DeepConv-DTI~\citep{Lee2019DeepConvDTIPO} uses a fully connected neural network to encode the ECFP4 drug fingerprint and a Convolutional Neural Network (CNN) along with a global max-pooling layer to extract features from protein sequences. Then the drug and protein features are concatenated and fed into a fully connected neural network for final prediction.

\noindent(4) GraphDTA~\citep{Nguyen2020GraphDTAPD} uses graph neural networks (GNNs) for the encoding of drug molecular graphs, and a CNN is used for the encoding of protein sequences. The derived vectors of the drug and protein representation are concatenated for interaction prediction.

\noindent(5) MolTrans~\citep{Huang2021MolTransMI} uses transformer architecture to encode drug and protein. Then a CNN-based interaction module is used to capture their interactions.

\noindent(6) DrugBAN~\citep{bai2023interpretable} use Graph Convolution Network (GCN)~\citep{DBLP:conf/iclr/KipfW17} and 1D CNN to encode drug and protein sequences. Then a bilinear attention network are adopted to learn pairwise local interactions between drug and protein. The resulting joint representation is decoded by a fully connected neural network.

\noindent{\textbf{Prompts}}

\noindent{\underline{Task Definition}}:
\textit{Definition: Drug target interaction prediction task (a binary classification task) for the \dataset{} dataset. If the given molecule and protein can interact with each other, indicate via "Yes". Otherwise, response via "No".} where \dataset{} is one of the three DTI datasets mentioned above.

\noindent{\underline{Instruction}}:
\textit{Now complete the following example - Input: Molecule: \bom{}\selfies{}\eom{} Protein: \bop{}\fasta{}\eop{} Output:}.

\noindent{\underline{Output}}:
\textit{Yes} for positive label or \textit{No} instead.

\subsubsection{Protein-protein Interaction Prediction}
\noindent{\textbf{Datasets}}

\noindent(1) Yeast~\citep{guo2008using} involves determining whether two yeast proteins interact or not. The negative pairs are derived from distinct subcellular locations. Following~\citep{xu2022peer}, the dataset is split and removed redundancy according to protein sequences similarity, which allows for the evaluation of generalization across dissimilar protein sequences.

\noindent(2) Human~\citep{pan2010large} involves determining whether two human proteins interact or not. It comprises positive protein pairs sourced from the Human Protein Reference Database (HPRD)~\citep{peri2003development} and negative pairs derived from different subcellular locations. The dataset splitting scheme is similar to that of Yeast PPI prediction with an $8:1:1$ ratio for train/validation/test. 

\noindent{\textbf{Baselines}}
The compared baselines are the same as the protein property prediction task in Section~\ref{sec:protein_property_detail}.

\noindent{\textbf{Prompts}}

\noindent{\underline{Task Definition}}:
\textit{Protein protein interaction prediction task (a binary classification task) for the \dataset{} dataset. If the given two yeast proteins (Protein\_A and Protein\_B) can interact with each other, indicate via "Yes". Otherwise, response via "No".} where \dataset{} is either \textit{yeast} or \textit{human}.

\noindent{\underline{Instruction}}:
\textit{Now complete the following example - Input: Protein\_A: \bop{}\fasta{}\eop{} Protein\_B: \bop{}\fasta{}\eop{} Output:}.

\noindent{\underline{Output}}:
\textit{Yes} for positive label or \textit{No} instead.

\subsection{Cross-modal Generation}
\subsubsection{Molecule Captioning}
\label{sec:mol2text_detail}
\noindent{\textbf{Datasets}}

\noindent We use ChEBI-20 dataset created by Text2mol~\citep{DBLP:conf/emnlp/EdwardsZJ21}, which consists of $33,010$ molecule-text pairs and $20$ means each text description has more than $20$ words.
The dataset is split into $8:1:1$ for train, validation, and test.

\noindent{\textbf{Baselines}}

\noindent(1) RNN~\citep{medsker2001recurrent} with $4$-layer bidirectional encoder is trained from scratch on ChEBI-20 dataset.

\noindent(2) Transformer~\citep{vaswani2017attention} containing $6$ encoder and decoder layers is trained from scratch on ChEBI-20 dataset.

\noindent(3) T5~\citep{raffel2020exploring} is directly fine-tuned on ChEBI-20 dataset from public checkpoints~\footnote{\url{https://github.com/google-research/text-to-text-transfer-transformer/blob/main/released_checkpoints.md\#t511}} with three different model sizes: small, base and large. Note that no molecule domain knowledge is introduced in the original T5 pre-training.

\noindent(4) MolT5~\citep{DBLP:conf/emnlp/EdwardsLRHCJ22} is jointly trained on molecule SMILES from ZINC-15 dataset~\citep{DBLP:journals/jcisd/SterlingI15} and general text from C4 dataset~\citep{raffel2020exploring} so that MolT5 has prior knowledge about these two domains. It also contains three different sizes: small, base and large. Then they are further fine-tuned on ChEBI-20 dataset.

\noindent(5) GPT-3.5-turbo~\citep{li2023empowering} is used by directly call OpenAI API without further fine-tuning. The input includes five parts as~\citet{li2023empowering}: role identification, task description, examples, output instruction, and user input prompt. The examples are retrieved by Morgan Fingerprint~\citep{DBLP:journals/jcisd/Butina99} similarity for molecule captioning task and by BM25~\citep{DBLP:journals/ftir/RobertsonZ09} for text-based molecule generation task.

\noindent(6) MolXPT~\cite{liu2023molxpt} is jointly trained on molecule SMILES from PubChem~\citep{kim2023pubchem}, biomedical text from PubMed~\citep{canese2013pubmed}, and ``wrapped'' text in which molecule names are replaced with corresponding SMILES.

\noindent{\textbf{Prompts}}

\noindent Different from the classification task in which the ground truth output is either \textit{Yes} or \textit{No}, the output for molecule captioning task is text sequence.

\noindent{\underline{Task Definition}}:
\textit{Definition: You are given a molecule SELFIES. Your job is to generate the molecule description in English that fits the molecule SELFIES.}

\noindent{\underline{Instruction}}:
\textit{Now complete the following example - Input: <bom>\selfies{}<eom> Output:}.

\noindent{\underline{Output}}:
\textit{\text{}}

\subsubsection{Text-based molecule generation}
This is the reverse task of molecule captioning.
The input is the text description of the desired molecule and the output is the corresponding molecule SELFIES.
The datasets and compared baselines are the same with molecule captioning in Section~\ref{sec:mol2text_detail} so will only provide the prompts here.

\noindent{\textbf{Prompts}}

\noindent{\underline{Task Definition}}:
\textit{Definition: You are given a molecule description in English. Your job is to generate the molecule SELFIES that fits the description.}

\noindent{\underline{Instruction}}:
\textit{Now complete the following example - Input: \text{} Output:}.

\noindent{\underline{Output}}:
\textit{<bom>\selfies{}<eom>}

\section{Case Study}
\label{sec:case_study}
\begin{figure*}[t]
    \centering
    % \vspace{-1.5cm}
    \includegraphics[width=\linewidth]{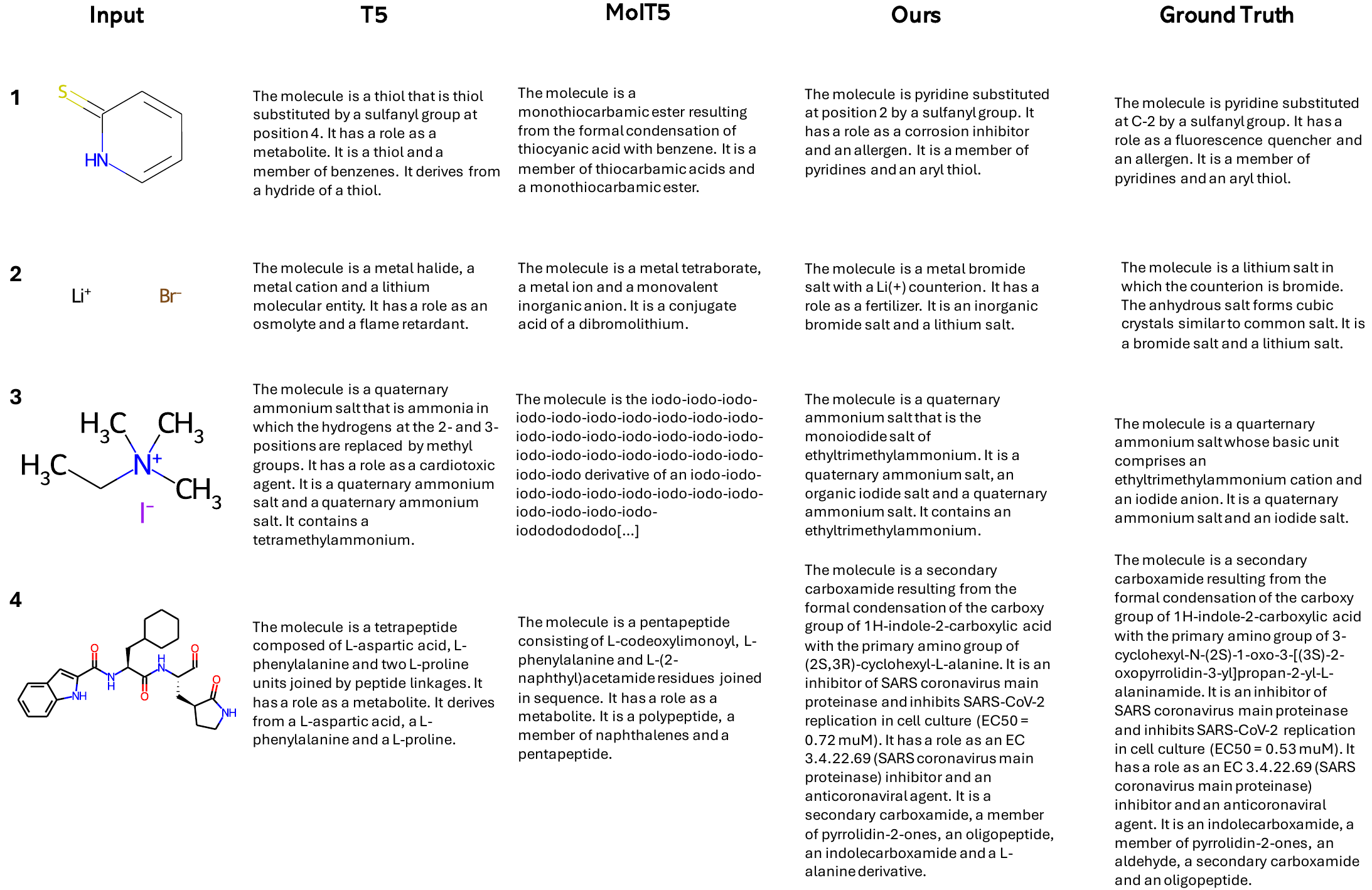}
    \caption{Molecule captioning cases.}
    \label{fig:case_mol2text}
\end{figure*}

\begin{figure*}[t]
    \centering
    % \vspace{-1.5cm}
    \includegraphics[width=\linewidth]{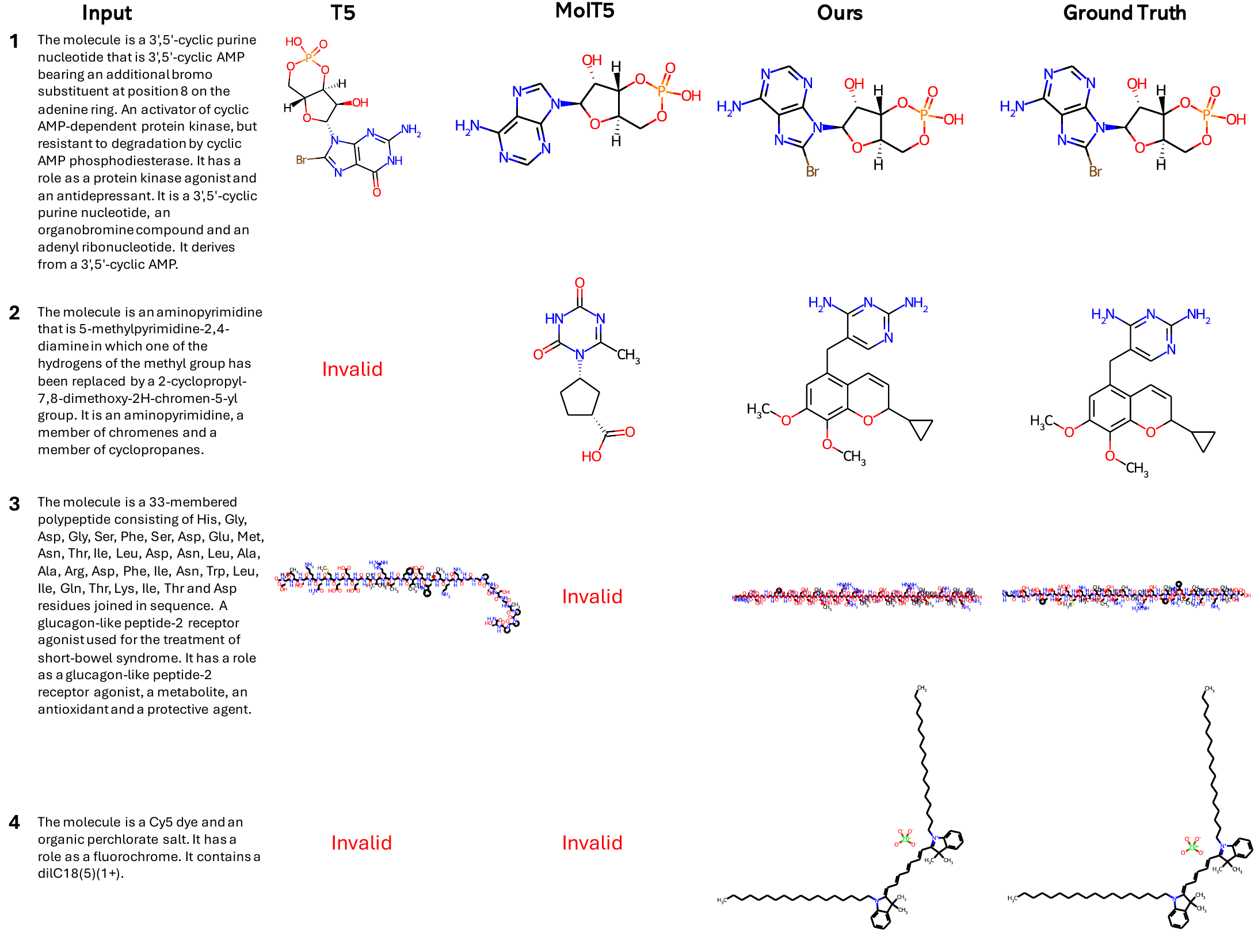}
    \caption{Text-based molecule generation cases.}
    \label{fig:case_text2mol}
\end{figure*}

In this section, we show several example outputs from different models in molecule captioning and text-based molecule generation tasks.
Figure~\ref{fig:case_mol2text} shows the cases for the molecule captioning task.
In example (1), the description of \method{} matches the ground truth best, successfully localizing the position of the substituent group and ``member of pyridines and an aryl thiol''.
In example (2), MolT5 mistakenly describes that the molecule contained boron, while \method's description is more accurate.
In example (3), while MolT5 generates repetitive output, \method{} and T5 generate semantically coherent output, and \method{}'s output matches better with ground truth.
For a complex molecule in example (4), the output of \method{} is more holistic and accurate.
Notably, only \method{} describes this molecule as an inhibitor of SARS coronavirus main proteinase, which may come from our integration with protein knowledge.
Figure~\ref{fig:case_text2mol} shows the cases for the text-based molecule generation task.
From the cases, we have several findings:
(i) \method{} is more likely to generate molecules that exactly match the ground truth.
(ii) By using SELFIES, \method{} will not generate invalid molecules, especially for complex and longer molecules shown in examples (3) and (4).
(iii) Some molecules are actually short proteins. Example (3) shows a molecule that is a 33-membered polypeptide, which consists of $33$ amino acid residues joined in sequence. Therefore, the boundary between proteins and molecules may not always be distinct, and leveraging information from both can provide reciprocal benefits.

\end{document}